# Clinical Evaluation of Medical Image Synthesis: A Case Study in Wireless Capsule Endoscopy

**Panagiota Gatoula[1], Dimitrios E. Diamantis[1], Anastasios Koulaouzidis[2,3], Cristina Carretero[4], Stefania Chetcuti-Zammit[5], Pablo Cortegoso Valdivia[6], Begoña González-Suárez[7,8], Alessandro Mussetto[9], John Plevris[10], Alexander Robertson[11], Bruno Rosa[12], Ervin Toth[13,14], & Dimitris K. Iakovidis[1]** ✉

Synthetic Data Generation (SDG) based on Artificial Intelligence (AI) can transform the way clinical medicine is delivered by overcoming privacy barriers that currently render clinical data sharing difficult. This is the key to accelerating the development of digital tools contributing to enhanced patient safety. Such tools include robust data-driven clinical decision support systems, and example-based digital training tools that will enable healthcare professionals to improve their diagnostic performance for enhanced patient safety. This study focuses on the clinical evaluation of medical SDG, with a proof-of-concept investigation on diagnosing Inflammatory Bowel Disease (IBD) using Wireless Capsule Endoscopy (WCE) images. Its scientific contributions include a) a novel protocol for the systematic Clinical Evaluation of Medical Image Synthesis (CEMIS); b) a novel variational autoencoder-based model for the generation of high-resolution synthetic WCE images; and c) a comprehensive evaluation of the synthetic images using the CEMIS protocol by 10 international WCE specialists, in terms of image quality, diversity, and realism, as well as their utility for clinical decision-making. The results show that TIDE-II generates clinically plausible, very realistic WCE images, of improved quality compared to relevant state-of-the-art generative models. Concludingly, CEMIS can serve as a reference for future research on medical image-generation techniques, while the adaptation/extension of the architecture of TIDE-II to other imaging domains can be promising.

## Introduction

Sharing medical data is crucial for developing data-driven digital tools, such as Artificial Intelligence (AI)-based Clinical Decision Support (CDS) systems, and digital twins for *in silico* medical device and drug testing[1]. Yet privacy regulations, such as the General Data Protection Regulation (GDPR)[2], pose significant challenges to this direction. AI-based Synthetic Data Generation (SDG) methods can essentially contribute to overcome these barriers, since they are capable of creating fully anonymized synthetic equivalents of the real medical data[1,3]. Therefore, their use promises a catalytic effect for the development of digital tools that could enhance patient safety, transforming the way clinical medicine is delivered. AI-based generative models for SDG, used to produce synthetic versions of original clinical data[1,6–8] include Generative Adversarial Networks (GANs)[4] and Variational Autoencoders (VAEs)[5]. Application domains include Computed Tomography (CT)[9], Magnetic Resonance Imaging (MRI)[10] and endoscopy[11,12]. To date, SDG has mainly been considered as a means of enhancing the performance of AI-based CDS systems. Such systems exploit knowledge that can be automatically extracted from large amounts of annotated retrospective data to assist clinicians in decision-making, *e.g.*, detecting or identifying suspicious lesions in medical images. Also, most of the current studies focus on assessing the impact of the use of synthetic data in the performance of the CDSs, without evaluating the clinical plausibility of the resulting images.

[1]Department of Computer Science and Biomedical Informatics, University of Thessaly, Greece, [2]Department of Clinical Research, University of Southern Denmark, Odense, Denmark, [3]Department of Gastroenterology, Pomeranian Medical University, Szczecin, Poland, [4]Department of Gastroenterology, Clínica Universidad de Navarra, Pamplona, Spain, [5]Gastroenterology Department, Mater Dei Hospital, MSD 2090, Malta, [6]Gastroenterology and Endoscopy Unit, University Hospital of Parma, University of Parma, Parma, Italy, [7]Department of Gastroenterology, Hospital Clínic i Provincial de Barcelona, Barcelona, Spain, [8]Centro de Investigación Biomédica en Red de Enfermedades Hepáticas y Digestivas, Madrid, Spain, [9]Department of Gastroenterology, S Maria della Croci Hosp, Ravenna, Italy, [10]The Centre for Liver and Digestive Disorders, Royal Infirmary of Edinburgh, Edinburgh, UK, [11]Department of Digestive Diseases, University Hospitals of Leicester NHS Trust, Leicester, UK, [12]Gastroenterology Department, Hospital da Senhora da Oliveira - Guimarães – Portugal, [13]Department of Gastroenterology, Skåne University Hospital, Lund University, Malmö, Sweden, [14]Department of Clinical Sciences, Lund University, Lund, Sweden. ✉ e-mail:diakovidis@uth.gr

Only a few studies have addressed synthetic image generation from a clinical viewpoint[11,13–18], most involving clinicians performing Visual Turing Tests (VTTs) that focus on discriminating between real and synthetic images. Previously, VTT procedures have been used to assess the plausibility of generated X-ray[16], CT[17], MRI[18] and retinal[15] images compared to the real ones. Fewer studies have clinically evaluated the results of endoscopic image generation methods with more thorough protocols[11,14]. *Yoon et al*[14] applied a GAN-based method[19] for the generation of colonoscopic images containing sessile serrated lesions, and four experts evaluated the quality of synthetic images using a three-point Likert scale in addition to performing a typical VTT. The more recent study of *Vats et al*[11] assessed the same generation methodology on a publicly available dataset, and then eight medical experts applied a threefold evaluation protocol to assess synthetic images produced. In addition to distinguishing synthetic images from real images, this protocol measured the difficulty medical experts have in performing the VTT and compared the degree of realism of the images with a sorting procedure. Also, it assessed the plausibility of endoscopic findings in synthetic image sequences by employing a four-point Likert scale of likelihood ranging between "*very unlikely*" to "*very likely*". However, not all the experts participated in the evaluation tasks performed.

Current evaluation protocols for synthetic image generation are limited to assessing only synthetic image quality or plausibility. More importantly, while existing protocols rely on the experience of medical experts to draw reliable results, they need to uncover the underlying factors influencing the clinicians' decision-making process. An additional aspect overlooked in related studies concerns the diagnostic accuracy of recognizing pathological conditions in the generated images from a clinical perspective. Although the concept of identifying pathological conditions in synthetic images is assessed through data-augmentation techniques applied in the training of AI classification algorithms, this aspect is not evaluated in the existing clinical evaluation protocols. It should be noted that the protocols used in previous studies are limited to evaluating only the methodologies proposed in these studies, and do not include any clinical comparisons with other image generation methodologies.

This study performs a proof-of-concept investigation on Inflammatory Bowel Disease (IBD) diagnosis based on evidence from Wireless Capsule Endoscopy (WCE) examinations. The selection of this domain is mainly motivated by the currently low diagnostic yield of WCE reported in clinical studies[20]. Synthetic WCE images can improve the diagnostic yield through lifelong clinical training and AI-based CDS systems' performance. WCE is a medical imaging modality that has become the prime choice for small-bowel examinations[21]. WCE uses a small swallowable capsule, with the size of a large vitamin pill, equipped with at least one colour camera, which, when swallowed, captures high-quality images of the gastrointestinal (GI) tract. Although WCE is effective, it requires significant time from experienced physicians to review the captured video, typically 60-90 minutes[22], which can lead to human error due to fatigue, lowering the overall accuracy of the examination[20]. To overcome that, CDSs have been used to alleviate this issue[23,24]. Yet, their performance is impacted by either the small number of images available for training or the class imbalance which characterizes WCE datasets. Most SDG methods used to battle this problem focus on abnormalities, such as polyps[11], characterized by homogeneity in texture and colour. A few SDG methodologies[25,26,13] have been proposed to replicate more complex conditions, such as IBD. Manifestations of IBD include various lesions, such as erythema, erosions, and ulcers, characterized by variations in colour, texture and anatomical structure. Despite these advancements, in the context of IBD, the quality of generated images has yet to be assessed through systematic clinical studies and has only been evaluated using typical VTT procedures[11,13,14].

This study addresses this gap by presenting a protocol for medical experts' systematic evaluation of synthetic images. The proposed protocol includes qualitative and quantitative aspects assessing the synthetic images' quality, variety, and plausibility regarding texture, anatomical structures, and diagnostic relevance. Furthermore, this study applies the proposed evaluation protocol to TIDE-II, which is the second version of our recently proposed deep learning architecture called 'This Intestine Does Not Exist' (TIDE)[13]. This revised version of the TIDE architecture is an enhanced generative model based on a variational autoencoder (VAE). It significantly improves upon the original TIDE[13] by generating high-resolution images that are four-fold larger than those produced in our previous study. This advancement allows for a broader variety of plausible images, showcasing the enhanced capability of the model. Unlike prior studies, this research incorporates a comprehensive qualitative evaluation involving an international group of 10 medical experts. These experts provided detailed assessments of the synthetic images, comparing various SDG methodologies and evaluating the images based on specific criteria of the proposed protocol, including texture, anatomical accuracy, and diagnostic difficulty.

In summary, this paper contributes to the field of medical image synthesis by proposing:

- A protocol for the Clinical Evaluation of Medical Image Synthesis (CEMIS) by physicians. CEMIS can serve as a reference for future research in medical image generation, setting a first standard for assessing synthetic data quality and its applicability in clinical practice.
- A novel VAE architecture, TIDE-II, which draws inspiration from architectural concepts of Visual Transformers (ViTs)[27] and can generate more realistic high-resolution images.
- A thorough clinical evaluation of the generated images by 10 experts with different levels of expertise demonstrates the utility of the proposed protocol and the effectiveness of TIDE-II over the relevant state-of-the-art architectures in the context of WCE for IBD lesion detection.

## Results

### Overview of Datasets

The clinical evaluation presented in this study was conducted on two publicly available WCE image datasets, namely, *KID 2*[28] and *Kvasir-Capsule datasets*[29]. KID 2 includes 2,371 fully annotated colour images with a resolution of 360×360 pixels, captured from the entire GI tract using the MiroCam® (IntroMedic Co., Seoul, Korea) capsule. It has a subset of 728 images illustrating healthy tissue of the small bowel and 593 images with pathological findings, out of which 227 represent inflammatory lesions. The Kvasir-Capsule dataset includes 47,238 weakly annotated colour images with a resolution of 336×336 pixels, captured using the Olympus ENDOCAPSULE 10 System with Olympus EC-S10 capsule. It has a subset of 34,338 images illustrating healthy tissue of the small-bowel, and 4,266 images with pathological findings, out of which 1,519 represent inflammatory lesions. The two described datasets were used to train TIDE-II and state-of-the-art generative models to produce synthetic normal and abnormal (with inflammatory lesions) WCE images.

### Medical experts

A group of 10 WCE specialists with 5-27 years of clinical experience reviewing WCE videos participated in clinically evaluating the synthetic images. Three had less than 10 years of experience, four had 10-20 years of experience, and three had over 20 years of experience. The CEMIS protocol involves 5 assessment procedures (*A1-A5*). In each procedure, the experts provided feedback while observing one or more images without any time limit. Once the experts had submitted their feedback, they were not able to reconsider or change their replies. All the experts participated in all assessment procedures.

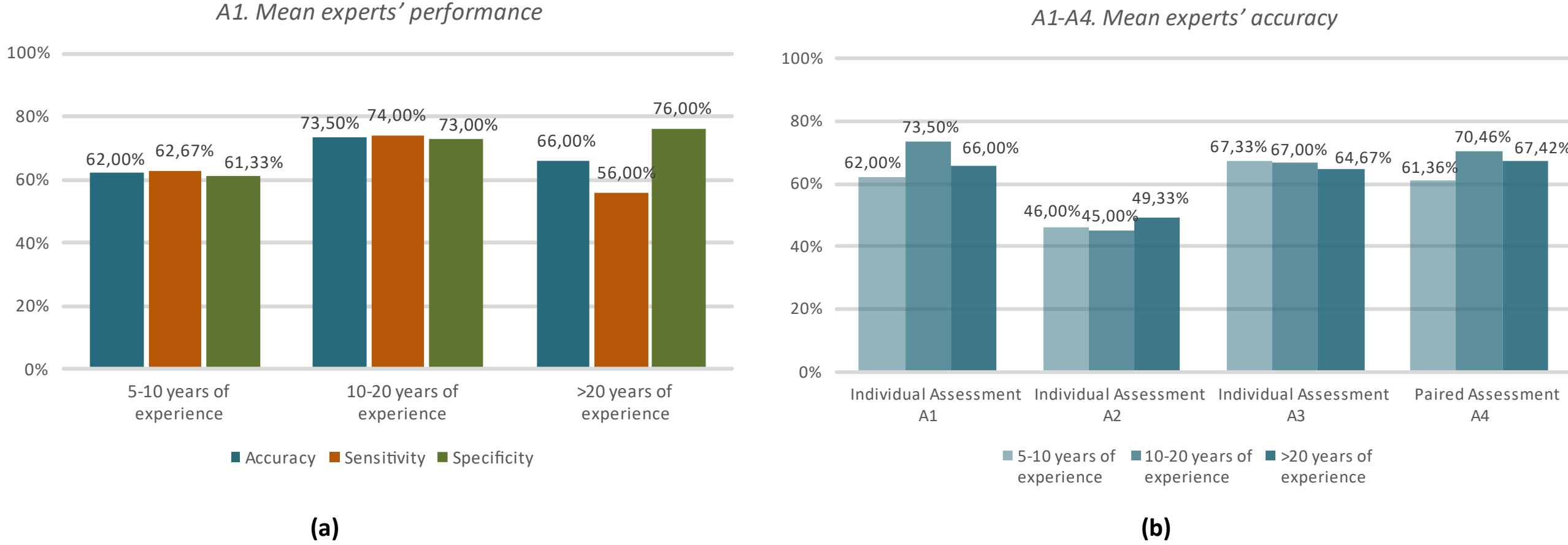

**Fig. 1 | Medical experts' performance in classification of images as real or synthetic, ordered by years of experience. (a)** Experts' performance in assessment procedure A1. **(b)** Experts' performance in assessment procedures A1-A4, in terms of accuracy. Sensitivity and specificity cannot be estimated since the tasks involve images belonging only in one category (A2 only synthetic, and A3 only real images).

**Table 1 | Mean medical experts' performance in discriminating between real and artificially synthesized images generated by TIDE-II model, per assessment procedure.**

| | Accuracy (%) | Sensitivity (%) | Specificity (%) |
|---|---|---|---|
| A1 Individual Assessment | 67.80 ± 10.85 | 65.20 ± 16.12 | 70.40 ± 21.35 |
| A2 Individual Assessment | 46.60 ± 6.54 | - | - |
| A3 Individual Assessment | 66.40 ± 11.88 | - | - |
| A4 Paired Assessment | 66.82 ± 16.04 | - | - |

### Individual Assessment of Real and Artificially Synthesized Images (A1)

A set of 50 WCE images (generated by TIDE-II) with a balanced distribution of samples in terms of type (real/synthetic), category (normal/abnormal), and origin (KID/Kvasir), was randomly selected and used for the first assessment procedure (*A1*) of the CEMIS protocol. In the first task (*A1.T1*), the experts classified the images as *real* or *synthetic*. Considering real images as positive and synthetic images as negative, the results of this task, in terms of mean accuracy, sensitivity and specificity, are summarized in Table 1. Further analysis of these results based on the experts' years of experience in WCE reviewing is presented in Fig. 1a. Experts with 10-20 years of experience performed best in accuracy and sensitivity. Supplementary Fig. 1 illustrates images correctly classified by all experts of the same experience groups.

The experts' predictions were evaluated using a Chi-Square test for goodness-of-fit, assuming that any differences in the rates of correct and incorrect predictions between real and synthetic images could be attributed to a random variation rather than a systematic effect. The overall probability of correct predictions for real images was 0.65, with a 95% confidence interval ranging from 0.59 to 0.71, whereas for synthetic images, it was 0.70, with a 95% confidence interval ranging from 0.64 to 0.76. The Chi-Square test yielded a $p$-value of $5.76\times10^{-14}$, which is significantly less than 0.05. However, for three out of ten medical experts, one from each group based on the years of experience, the $p$-values were greater than 0.05, indicating no statistically significant differences in frequencies of correct and incorrect predictions between the real and synthetic images. This suggests that the images evaluated in this procedure were challenging for 30% of the experts.

In the second task (*A1.T2*), the clinicians rated the difficulty level regarding their choice in the previous task, *i.e.*, if the image presented was real or synthetic. The assessments of the experts are summarized in Figs. 2a and 2b. The real images (Fig. 2a) were considered *difficult* to classify by the least experienced and the most experienced experts, and as *easy* by the experts with 10-20 years of experience. Synthetic images were regarded as *easy* by experts with less than 10 years of experience, as *difficult* for those with 10-20 years, and as *neutral* by experts with over 20 years of experience.

In the third task (*A1.T3*), the experts explained the reasons behind their decision on whether the images evaluated were real or synthetic. Their assessments are summarized in Fig. 2c. C*olor* and *texture* affected most of the experts' decisions. Furthermore, both real and synthetic images were associated with the *existence of artifacts or luminal content* and with the *existence of unrealistic anatomical structures* in their content.

In the fourth task (*A1.T4*), the experts classified the images into normal or abnormal (containing inflammatory lesions). Considering abnormal images as positive, and normal images as negative, the results of this task in terms of accuracy, sensitivity and specificity are summarized in Table 2. It can be noticed that the classification was more accurate in the case of the real images, and that the lower accuracy observed in the case of the synthetic images was mainly due to the higher false-negative classifications. The false negatives could be attributed to the normal synthetic images containing more conspicuous structures than the real ones. The classification performance of the experts on real and synthetic images, accordingly, with respect to their experience, is provided in Figs. 2d and 2e. These figures validate that the overall observations are consistent across the different experience groups. In the fifth task (*A1.T5*), the medical experts evaluated the displayed image quality. The quality of most of the real images (Fig. 2f) was rated as *acceptable,* and the quality of most of the synthetic images (Fig. 2g) was rated as *slightly acceptable* by all the experts regardless their experience.

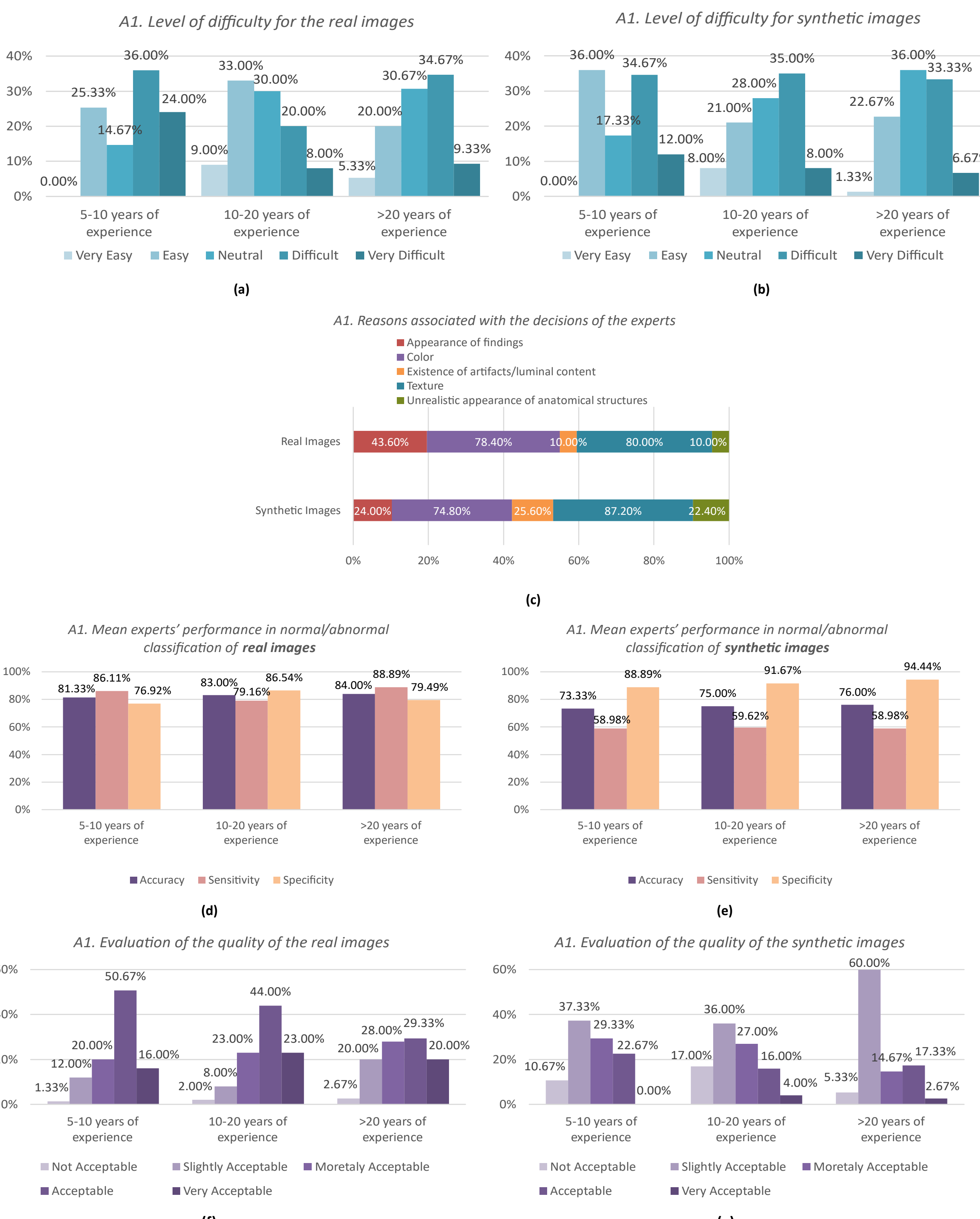

**Fig. 2 | Results obtained from the individual assessment of real and synthetic images generated by TIDE-II.** Medical experts' responses regarding the difficulty in classifying images as: **(a)** real, or **(b)** synthetic. **(c)** Reasons associated with the experts' decisions in classifying images as real or synthetic**.** Expert's performance in classifying the images as normal or abnormal (considering abnormal images as positive) for **(d)** real, and **(e)** synthetic images. Experts' assessment regarding the quality of the images for **(f)** real, and **(g)** synthetic images.

Supplementary Figs. 2 and 3 illustrate representative results from both real and synthetic images, rated as *acceptable* and *slightly acceptable* by the expert groups and the experts as a whole. Observing the synthetic images rated as *acceptable* and *slightly acceptable*, the experts converged to that both preserve the clinical features of real images, with ulcers (2b, 3a, and 3c), erosions (2a), and normal mucosa (2d, 3c, 3d) represented best.

### Individual Assessment of Synthetic Images (A2)

In the second assessment procedure (*A2*) of the CEMIS protocol, a total of 50 synthetic WCE images (generated by TIDE-II) with a balanced distribution of samples in terms of category (normal/abnormal), and origin (KID/Kvasir), was randomly selected for evaluation by the medical experts. In the first task (*A2.T1*), the experts classified the images presented as *real* or *synthetic*. The mean accuracy of the predictions performed by the clinicians was 46.60% ± 6.54% (Table 1). This indicates that 53.40% of the images generated by TIDE-II from both datasets were incorrectly perceived as real, whereas 46.60% were correctly identified as synthetic. An overview of the classification performance of the medical experts in procedure *A2* in comparison with their performance in *A1* upon their experience is included in Fig. 1b, where it can be noticed that it is significantly lower. Supplementary Fig. 4 illustrates synthetic images misperceived as genuine by all the medical experts of the same experience groups.

Considering that the average accuracy obtained by the medical experts was below 50.00% in this assessment, a Binomial test was performed to determine whether the observed predictions were statistically significantly lower than expected under the assumption of no systematic advantage attributed to the artificial origin of synthetic images. For the 10 medical experts, the test yielded a $p$-value of 0.07 with a 95% confidence interval ranging from 0.42 to 0.51. Moreover, all the $p$-values of the clinicians were more significant than 0.05 (ranging between 0.84 and 0.06). These results suggest that there is no statistically significant evidence to indicate that the synthetic images produced by the TIDE-II model are biased towards being perceived as artificial.

Figure 3a summarizes the experts' responses concerning the difficulty level in predicting if the images displayed were real or synthetic (task *A2.T2*). These results indicate that TIDE-II generates images that are challenging to be distinguished as synthetic. Synthetic images generated by TIDE-II were considered *difficult* to identify as synthetic by experts with less than 20 years of experience. The opinion of the most experienced clinicians (>20 years of experience) was more divergent, rating them as *easy* and *difficult* at similar rates. Figure 3b presents the reasons (task *A2.T3*) behind the medical experts' decision regarding the first task of this assessment. It can be noticed that in total, 17.80% of the synthetic images were associated with the *unrealistic appearance of anatomical structures*, and this was the main characteristic that drove the experts to correctly identify the synthetic images, since it appears in 38.20% of the correct predictions and 0.00% of the incorrect predictions.

Table 2 summarizes the results of the experts in classifying synthetic images as *normal* or *abnormal* (task *A2.T4*). It can be noticed that the mean sensitivity estimated for the images generated based on the KID dataset was significantly higher than that of the Kvasir dataset. In contrast, in the case of Kvasir-based synthetic images, the mean specificity was considerably higher than that of KID-based synthetic images. This implies that the KID-based abnormal synthetic images can be better recognized as abnormal, and the normal Kvasir-based synthetic images can be better recognized as normal. The comparison of the overall performance of the experts in relation to their years of experience is presented in Fig. 3c, where it can be noticed that the most experienced clinicians were more accurate in this task.

**Table 2 | Mean accuracy, sensitivity and specificity for the classification of the real and the synthetic images generated by TIDE-II into normal and abnormal categories.**

| | Accuracy (%) | Sensitivity (%) | Specificity (%) |
|---|---|---|---|
| **A1 Individual Assessment of Real and Artificially Synthesized Images** | | | |
| Real Images | 82.80 ± 4.64 | 84.17 ± 9.98 | 81.54 ± 9.03 |
| Synthetic Images | 74.80 ± 3.79 | 59.23 ± 10.29 | 91.67 ± 8.42 |
| In Total | 78.80 ± 3.29 | 71.20 ± 8.18 | 86.40 ± 9.83 |
| **A2 Individual Assessment of Synthetic Images** | | | |
| Real Images | - | - | - |
| Synthetic Images | | | |
| KID dataset | 74.80 ± 4.64 | 82.50 ± 15.44 | 67.69 ± 16.54 |
| Kvasir dataset | 70.80 ± 4.64 | 45.84 ± 11.95 | 93.85 ± 4.86 |
| In Total | 72.80 ± 3.68 | 64.16 ± 11.98 | 80.77 ± 9.42 |
| **A3 Individual Assessment of Real Images** | | | |
| Real Images | | | |
| KID dataset | 90.40 ± 5.40 | 86.67 ± 11.25 | 93.85 ± 5.76 |
| Kvasir dataset | 89.20 ± 8.65 | 78.33 ± 17.21 | 99.23 ± 0.43 |
| Synthetic Images | - | - | - |
| In Total | 89.80 ± 5.61 | 82.50 ± 11.75 | 96.54 ± 3.40 |
| **A4 Paired Image Assessment** | | | |
| Real Images | 80.23 ± 4.42 | 66.67 ± 15.49 | 89.62 ± 10.10 |
| KID dataset | 79.17 ± 7.08 | 70.00 ± 18.18 | 84.67 ± 15.09 |
| Kvasir dataset | 81.50 ± 4.74 | 63.33 ± 13.91 | 96.36 ± 3.36 |
| Synthetic Images | 82.04 ±4.60 | 69.04 ± 13.88 | 93.91 ± 5.72 |
| KID dataset | 80.00 ± 6.45 | 70.00 ± 16.76 | 90.00 ± 9.91 |
| Kvasir dataset | 84.50 ± 6.43 | 67.78 ± 17.72 | 98.18 ± 1.83 |

The results from the quality evaluation of the images generated by TIDE-II (task *A2.T5*) are presented in Fig. 3d. It can be observed that synthetic images were rated as *acceptable* by the experts with 5-10 years of experience and as *moderately acceptable* by the experts with 10-20 years of experience. However, the clinicians with over 20 years of experience rated the quality of synthetic images as *slightly acceptable* at a percentage of 30.00%. In contrast, the quality of most synthetic images (42.00%) was rated as *acceptable* and *very acceptable*.

### Individual Assessment of Real Images (A3)

In the third assessment procedure (*A3*) of the CEMIS protocol, the medical experts evaluated solely real WCE images. A set of 50 real WCE images with a balanced distribution of samples in terms of category (normal/abnormal) and origin (KID/Kvasir) was randomly selected. As in the previous assessment procedures, in the first task (*A3.T1*), the clinicians decided whether the images presented were *real* or *synthetic*. The mean accuracy was 66.40 ± 11.88% (Table 1). An overview of the classification performance of the medical experts on both datasets with respect to the years of experience is included in Fig. 1b.

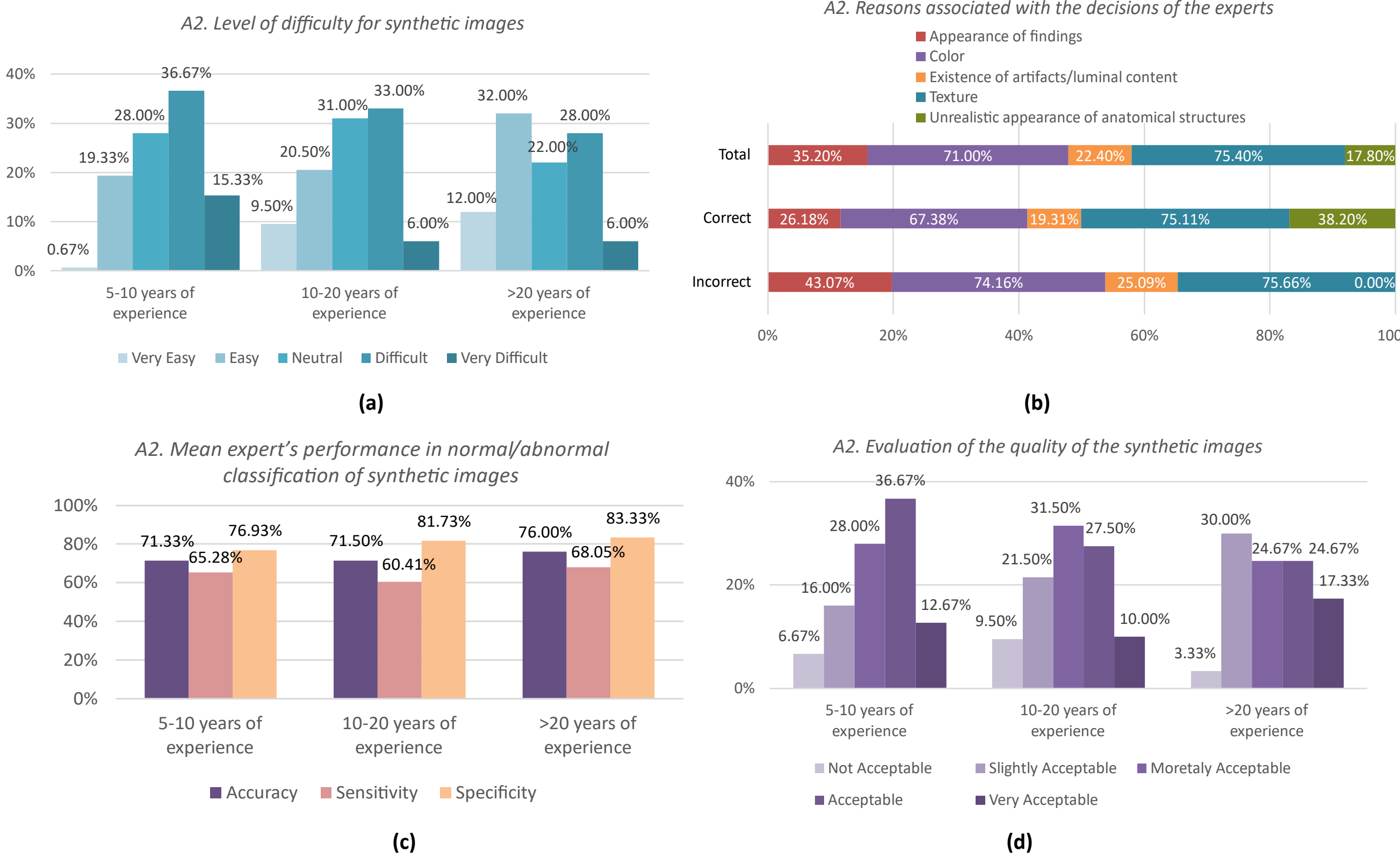

**Fig. 3 | Results from individual image assessment of synthetic images generated by TIDE-II. (a)** Medical experts' responses regarding the difficulty in classifying the images as real or synthetic. **(b)** Reasons associated with the experts' decisions in classifying the images as synthetic (correct) or real (incorrect). **(c)** Experts' mean performance in classifying the synthetic images as normal or abnormal. **(d)** Results of the synthetic image quality evaluation.

A two-sided Binomial test was applied to evaluate whether the predictions were statistically significant from what would be expected under the assumption of no systematic advantage attributed to the appearance of real images. The test yielded a *p*-value of $1.90\times10^{-13}$ for the ten experts, which is significantly lower than 0.05, with a 95% confidence interval ranging from 0.62 to 0.70. For four out of ten experts (one expert with 5-10 years of experience, two with 10-20 years of experience, and one with more than 20 years of experience), the test yielded a *p*-value greater than 0.05. These results suggest that there is statistically significant evidence that real images have a distinguishing advantage, but this effect varies among different experts based on their experience.

Figure 4a shows the level of difficulty, based on the experts' opinions, in predicting whether the images displayed were real or synthetic (task *A3.T2*). In most cases, the experts with 5-10 years of experience characterized distinguishing real images as a task of *neutral* difficulty. It is shown that the overall opinion of the more experienced groups of experts tends to have a more balanced distribution with similar levels of difficulty for *easy* and *difficult* categories. These results indicate that real WCE images exhibit diverse perceptions for their difficulty level.

Figure 4b analyzes the reasons behind the experts' decision regarding whether the images presented were real or synthetic (task *A3.T3*). Although all the images examined were real, there were images (11.40%) in which the experts characterized some anatomical structures as having an *unrealistic appearance*. The diagram shows that this was the main reason for the experts' incorrect classification of the real images (33.93%) as synthetic.

Table 2 summarizes the results of the experts' classification of real images as normal or abnormal (task *A3.T4*), and Fig. 4c presents the performance of experts in relation to the years of their experience. It can be noticed that the experts performed better overall with the real images than with the synthetic ones. Figure 4d summarizes the results of the quality of the real images (task *A3.T5*), where it can be observed that it has been mainly considered as *acceptable* by the different groups of experts.

### Paired Image Assessment (A4)

This assessment procedure involved pairwise evaluation of the synthetic images generated by TIDE-II with the real images. Two new subsets of 50 synthetic and 50 real images, with a balanced distribution of samples in terms of type (real/synthetic), category (normal/abnormal), and origin (KID/Kvasir), were randomly selected and provided for evaluation to the experts as pairs of images composed by one synthetic and one real image.

In the first task (*A4.T1*), the experts identified which was the real image of each pair displayed. The mean accuracy obtained based on the clinicians' predictions was 66.82±16.04%. Aiming to determine the statistical significance of this result, a two-sided Binomial test was performed. The results obtained show that, for all experts, the overall predictions were statistically significant (*p*-value<0.05). In addition, for 50% of the experts (two experts with 5-10 years of experience, two with 10-20 years of experience, and one with more than 20 years of experience), the discrimination against real images was quite challenging (*p*-value>0.05). This suggests that for half of the experts, the synthetic images generated by TIDE-II were difficult to distinguish from the real images when placed side-by-side.

In the second task (*A4.T2*), the clinicians specified the difficulty level concerning the assessments submitted in the previous task. Figure 5a presents the assessments collected. It can be noticed that, in total, the experts from all experience groups rated more than 65% of the image pairs examined as *difficult* and *very difficult*. This indicates that TIDE-II synthesizes images that are hard to distinguish next to the real WCE images. Supplementary Fig. 5 presents the image pairs that were characterized as the most challenging ones.

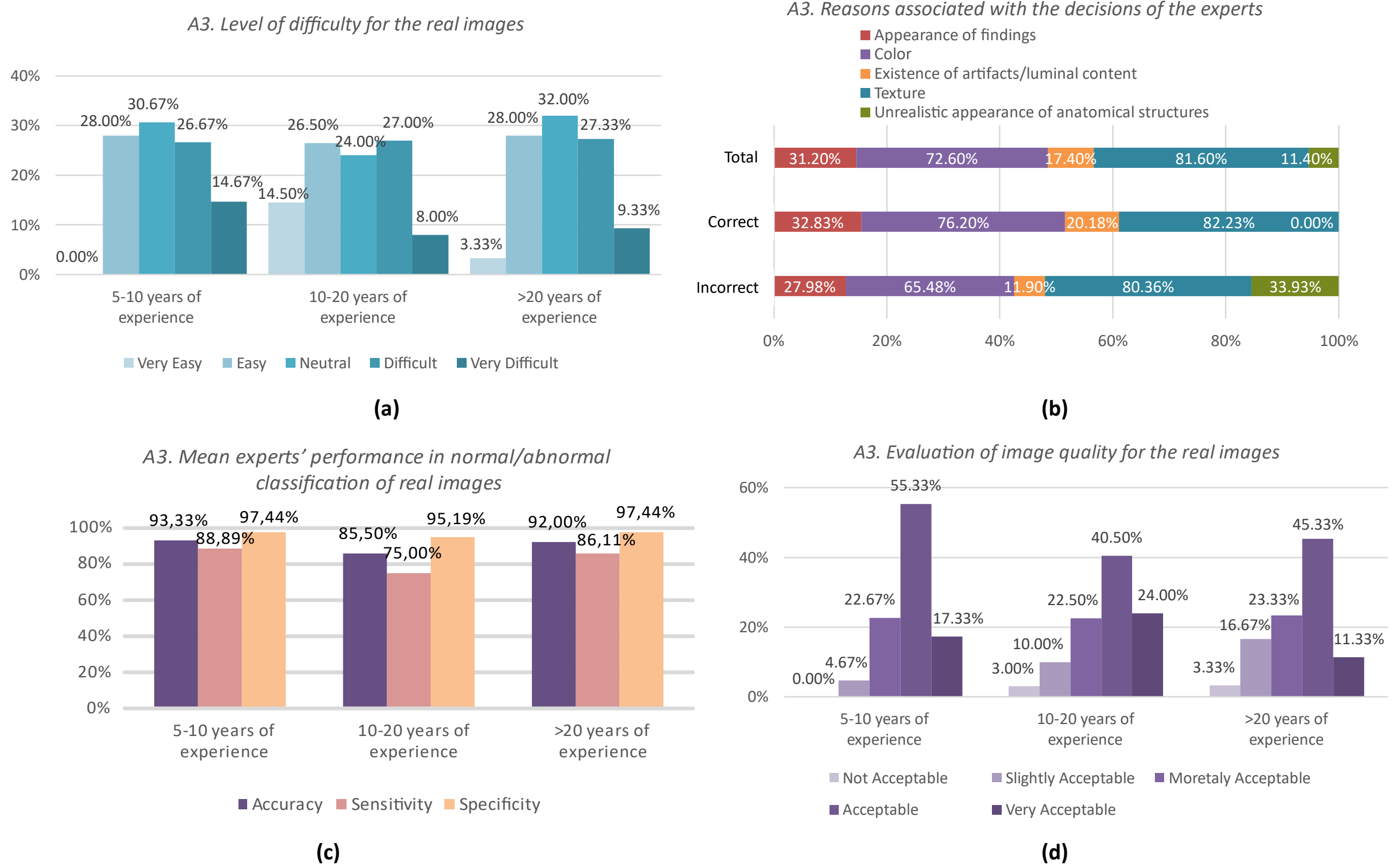

**Fig. 4 | Results from individual image assessment of real images. (a)** Medical experts' responses regarding the difficulty in classifying the images as real or synthetic. **(b)** Reasons associated with the experts' decisions in classifying the images as real (correct) or synthetic (incorrect). **(c)** Experts' mean performance in classifying the real images as normal or abnormal. **(d)** Results of the real image quality evaluation.

In the third task (*A4.T3*), the experts' motivation in deciding which was the real image in each pair was investigated. An overview of the assessments submitted is presented in Fig. 4b. The experts attributed their choice mainly to the *texture* and *color* properties. The rates associated with the *realistic appearance of findings* and *anatomical structures* to the content of the images were similar to each other. In contrast, the percentage of images related to the *absence of artifacts* was significantly lower. Furthermore, it is worth noting that in cases where the synthetic images were misperceived as real (incorrect predictions), it was attributed to the *absence of artifacts* and the *existence of findings with a realistic appearance* at rates of 10.27% and 32.88%, respectively, in contrary to the cases where the predictions were correct and the above reasons were rated at 8.84% and 20.07% respectively. This implies that the images synthesized by TIDE-II tend to be considered equivalent to the real ones for similar reasons concerning their clinical appearance.

Moreover, the clinicians classified the images of each pair as normal or abnormal when representing pathological findings (tasks *A4.T4a, A4.T4b*). At the same time, the origin, *i.e.*, real or synthetic, of the images remained undisclosed. Abnormal images were regarded as positive predictions, whereas normal images were considered as negative predictions. The respective results are included in Table 2, where it can be noticed that the experts' performance was comparable for the real and the synthetic images. A comparison of the total performance of the clinicians in relation to the years of their experience is provided in Figs. 5c (real images) and 5d (synthetic images). The classification results indicate that TIDE-II generates normal and abnormal images with distinct features that contribute to the discrimination of pathological and non-pathological conditions when compared side-by-side with real images.

The last tasks (*A4.T4a, A4.T4b*) of paired image assessment evaluated the quality of images included in each pair. Figures 5e and 5f summarize the results of the evaluation. The quality of real images (Fig. 5e) and synthetic images quality (Fig. 5f) were mainly considered *acceptable* by experts with 5-10 years and over 20 years of experience, respectively. The experts having 10-20 years of experience evaluated the quality of real and synthetic images as *very acceptable* in most cases.

### Evaluation of Diversity and Realism (A5)

In the last part of the CEMIS protocol, the experts evaluated the realism and the diversity of both real and synthetic images that were provided to them in different image groups. This was the fifth assessment procedure (*A5*), which included 600 randomly selected images composed of two parts: a part with 300 synthetic images and a part with 300 real images. Each part had a balanced distribution of samples in terms of category (normal/abnormal) and origin (KID/Kvasir). The part with the synthetic images was composed of 6 different subsets of 50 images generated using state-of-the-art SDG methods, three GAN-based (StyleGANv2, CycleGAN, TS-GAN), and three VAE-based (EndoVAE, TIDE, and TIDE-II). The images were organized into 60 image groups, 30 groups with real images and 30 with synthetic ones, balanced in terms of category and origin. Each group was composed of

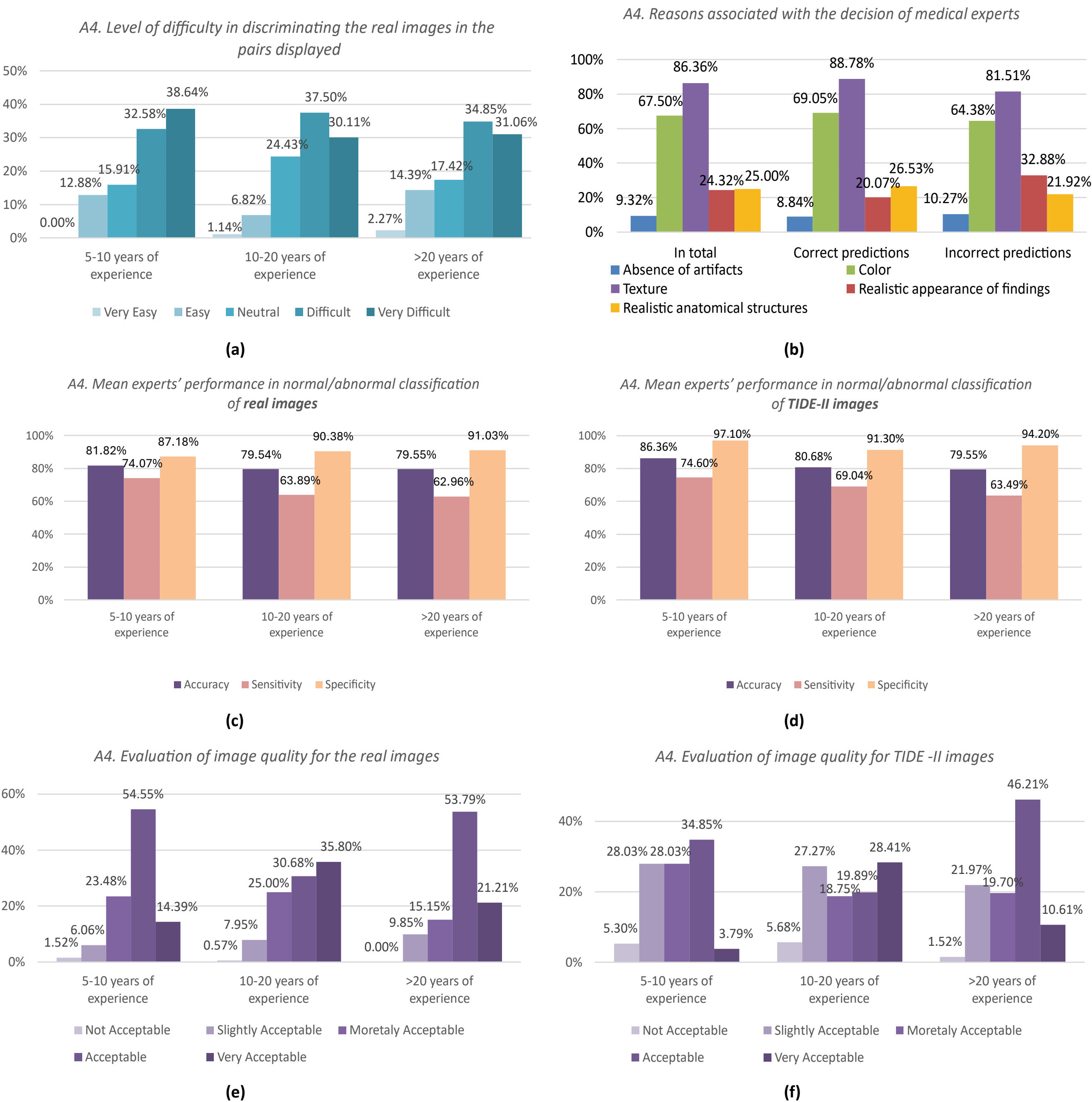

**Fig. 5 | Results of Paired image Assessment.** In paired image assessment medical experts evaluated pairs of images, each consisting of one real and one synthetic image generated by TIDE-II model. This information was disclosed to the medical experts at the beginning of the assessment. **(a)** Clinical experts' difficulty in deciding which of the two images included in each pair was the real one. **(b)** Reasons justifying the experts' decision for the image selected as real. **(c)** and **(d)** Experts' performance in classifying the images of each pair as normal and abnormal; **(c)** refers to real images and **(d)** to synthetic images. **(d)** and **(e)** Results of the quality image evaluation; **(d)** refers to real and **(e)** to synthetic images.

10 randomly selected images. The origin of the image groups presented to the experts remained undisclosed during this assessment.

Figure 6 provides a comparison of the results obtained from the evaluation of the real WCE images and synthetic images generated by the proposed TIDE-II model and its predecessor, TIDE. Regarding realism (Fig. 6a), most images from the real datasets were rated as *realistic*. Most synthetic images from TIDE were rated as *slightly realistic*, whereas most synthetic images from TIDE-II were rated as *moderately realistic*. Regarding the diversity (Fig. 6b), most real images were rated as *not diverse,* whereas the synthetic images generated by TIDE and TIDE-II were rated as *slightly diverse*.

Additionally, according to the experts, the TIDE model does not synthesize *very diverse* image collections. It can be noticed that TIDE-II exceeded its predecessor model in terms of realism and diversity. Supplementary Figs. 6 and 7 present a comparison of representative image collections generated by the TIDE and TIDE-II models, rated by the experts as the most realistic and diverse, respectively. An overall observation made by the experts on these image collections is that those generated by TIDE-II have a higher definition and they are more realistic.

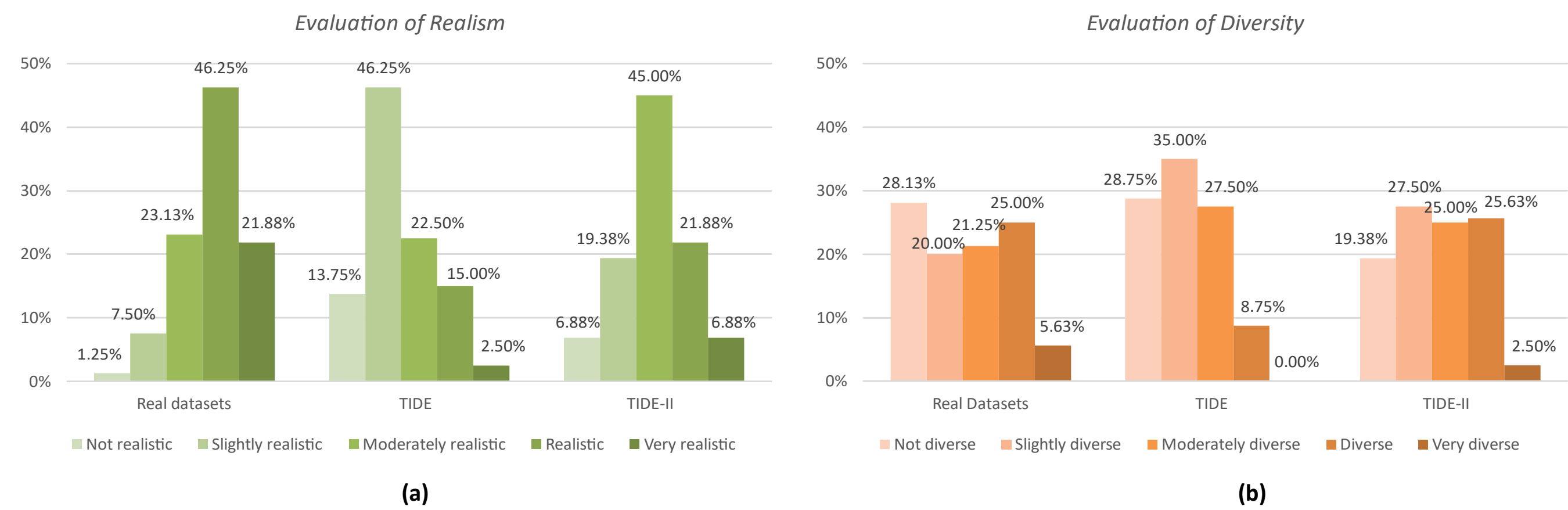

**Fig. 6 | Comparative results from the evaluation of diversity and realism.** In A5 assessment procedure the medical experts evaluated groups consisting of real or synthetic images produced by various synthesis methods in terms of realism and diversity. Experts' results of **(a)** realism and **(b)** diversity assessment, between the synthetic images generated by TIDE and TIDE-II models and the real images.

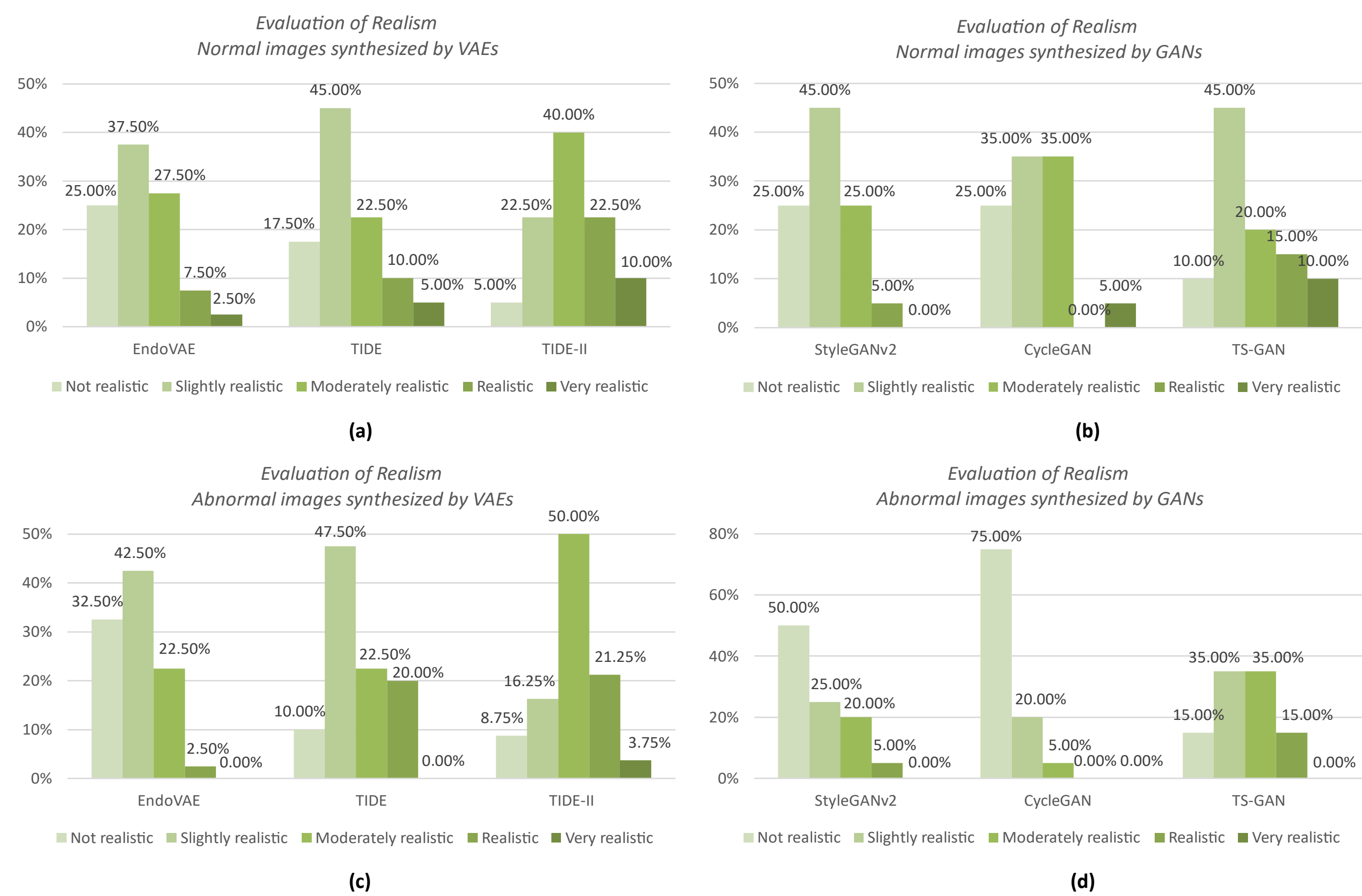

**Fig. 7 | Comparison of realism for state-of-art generative methodologies for endoscopic synthesis containing inflammation conditions**. In A5 assessment the medical experts evaluated the realism of synthetic images produced by various state-of-the-art synthesis methodologies including VAEs and GANs. Experts' results for synthetic normal images generated by **(a)** VAE models (EndoVAE, TIDE, TIDE-II) and **(b)** GAN models (StyleGANv2, CycleGAN, TS-GAN). Experts' results for synthetic abnormal images generated by **(c)** VAE models (EndoVAE, TIDE, TIDE-II) and **(d)** GAN models (StyleGANv2, CycleGAN, TS-GAN). Overall, the generation performance of TIDE-II model exceeds relative state-of-art methodologies in terms of realism for synthetic normal and abnormal images containing inflammation lesions.

A detailed analysis of the realism between the synthetic images generated from various state-of-the-art generative models, including VAE-based and GAN-based methodologies, is presented in Fig. 7. Most synthetic normal images generated by EndoVAE and TIDE, were evaluated as *slightly realistic* (Fig. 7a). Moreover, only 7.50% of synthetic normal images from EndoVAE and 10.00% from TIDE were evaluated as *realistic*, while the corresponding percentage in TIDE-II exceeded 20.00%. More than 50% of the normal images generated by GANs were considered as *not realistic* and *slightly realistic* (Fig. 7b). Regarding synthetic abnormal images, TIDE and EndoVAE were considered to synthesize *slightly realistic* images (Fig. 7c). Abnormal images from StyleGANv2 and CycleGAN were regarded as *not realistic*, whereas only 15.00% of the TS-GAN abnormal images were regarded as *realistic* (Fig. 7d). It can be observed that none of the CycleGAN normal and abnormal images were rated as *realistic*. Additionally, according to the opinion of the experts, only TIDE-II generates *very realistic* abnormal images among all state-of-the-art methodologies.

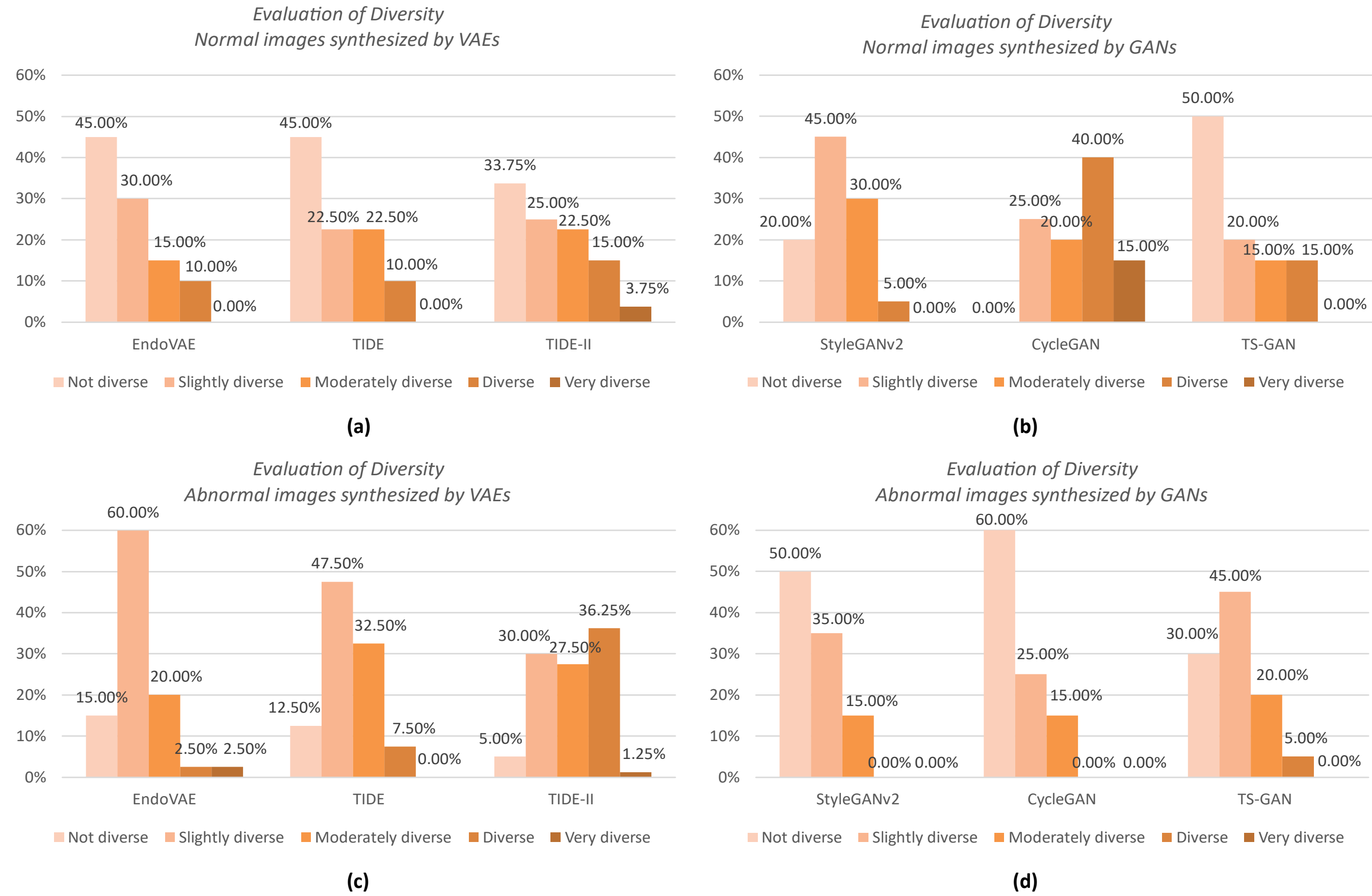

**Fig. 8 | Comparison of diversity for state-of-art generative methodologies for endoscopic synthesis containing inflammation conditions.** In A5 assessment the medical experts evaluated the diversity of synthetic images produced by various state-of-the-art synthesis methodologies including VAEs and GANs. Experts' results for synthetic normal images generated by (**a**) VAE models (EndoVAE, TIDE, TIDE-II) and (**b**) GAN models (StyleGANv2, CycleGAN, TS-GAN). Experts' results for synthetic abnormal images generated by (**c**) VAE models (EndoVAE, TIDE, TIDE-II) and (**d**) GAN models (StyleGANv2, CycleGAN, TS-GAN).

Figure 8 presents a detailed analysis of diversity assessment between synthetic images generated by VAE-based and GAN-based models. It is shown that TIDE-II generates less normal images rated as *not diverse* compared to previous VAEs (Fig. 8a). Additionally, neither EndoVAE nor TIDE normal images were rated as *very diverse*. Normal images from TS-GAN and StyleGANv2 were considered as *not diverse* and *slightly diverse*, accordingly (Fig. 8b). Normal images from CycleGAN were considered as *diverse*.

Regarding abnormal synthesis, images from EndoVAE and TIDE models were assessed as *slightly diverse* (Fig. 8c). In contrast, the images synthesized by TIDE-II were better rated as *diverse*, with a difference of 28.75% and 33.75% over EndoVAE and TIDE, respectively. Abnormal images from StyleGANv2 and CycleGAN were rated as *not diverse*, whereas those from TS-GAN were rated as *slightly diverse* (Fig. 8d). However, neither StyleGAN nor CycleGAN produced *diverse* abnormal images. In total, according to the opinion of medical experts, StyleGANv2 and TS-GAN generated neither normal nor abnormal *diverse* images.

## Discussion

Medical image synthesis has stimulated significant research interest; however, the evaluation of such images has yet to be systematically studied, primarily due to the lack of domain-specific metrics that access both the quality and clinical relevance of the synthetic images. To this end, in this study, we introduced the CEMIS protocol for evaluating synthetic medical images, which is applied in the context of WCE image generation for IBD diagnosis. Furthermore, we introduced a novel VAE-based architecture, named TIDE-II, for synthetic WCE image generation. We evaluated its performance compared to state-of-the-art WCE image synthesis methodologies using the proposed CEMIS protocol.

The results of the proposed CEMIS protocol indicate that synthetic images produced by TIDE-II are hard to distinguish from real ones. This difficulty was particularly evident in the individual image assessment (A1), where all experts consistently described the task as challenging (Figs. 2a and 2b). This difficulty was further corroborated by the evaluations conducted in assessments A2 and A3, where TIDE-II images were consistently more challenging to identify as synthetic than as real images. Furthermore, the paired assessment (A4), where real and synthetic images were presented side-by-side, was also very challenging, with experts across all groups finding it difficult to distinguish the synthetic images generated by TIDE-II from the real ones (Fig. 5a).

Across all assessments (A1-A4), we observed that experts primarily relied on low-level properties like color and texture, rather than higher-level features such as anatomical findings. Although in assessment A1, only the synthetic images were associated with unrealistic anatomical structures and artifacts (Fig. 2c), this was not validated in the individual assessment of solely real images (A3), which shows that real images also exhibited artifacts and unrealistic structures present in 17.00% and 11.00% of the cases, respectively. The paired assessment (A4) further revealed that both real and synthetic images exhibited similar levels of artifacts when compared side-by-side, with TIDE-II images occasionally displaying more realistic features (Fig. 5b). These findings indicate that even real datasets include images with inherent flaws or characteristics that can be misperceived and affect the experts' judgement to characterize them as non-plausible. The image synthesis models subsequently replicate these, and since they are not so common in the real training datasets, their synthesis can be less accurate. Regarding the artifacts, Supplementary Fig. 8 illustrates representative cases of artifacts detected in both synthetic and real images. Observing these images, the experts noted that the representation of the non-mucosal areas, such as debris and bubbles, or the areas under them, is not sufficiently realistic, which makes them distinguishable from the real ones, *e.g.* the whitish area in the center of Fig. 8a, the mucosa behind the debris on the left part of Fig. 8b, and the debris in the lower right part of Fig. 8d.

In both real and TIDE-II synthetic images, experts were more accurate in the identification of normal than abnormal images. This can be attributed to the nature of IBD findings. Inflammatory lesions include a variety of findings of different size such as erosions, erythema, ulcers etc. Additionally, the appearance of these lesions may have colour gradations and vary in shape as inflammation conditions can be flat or excavated.

An additional aspect examined in the proposed clinical protocol was the quality of both real and synthetic images. The results demonstrated that the quality of real images was mainly evaluated as *acceptable* in the assessments of our protocol (A1, A3, A4). More specifically, in individual assessment of solely TIDE-II images (A2), most of the synthetic images were rated as *acceptable* and *very acceptable* (Fig. 3d), and in paired assessment (A4), TIDE-II images were considered as *acceptable* or *very acceptable*. Only in individual image assessment (A1), where both real and synthetic images were present, the quality of TIDE-II images was rated as *slightly acceptable* (Fig. 2g).

Unlike relevant WCE image synthesis assessment protocols, CEMIS allowed a comprehensive comparative assessment of state-of-art GAN-based and VAE-based endoscopic synthesis methodologies by evaluating both the realism and diversity of the synthetic images in relation to real ones (A5). The results demonstrate that TIDE-II outperforms previous models, generating more realistic and diverse images (Figs. 7 and 8). Although the images generated by CycleGAN were noted for their higher diversity, their plausibility was significantly lower than that of the images generated by VAE-based methodologies. This discrepancy arises because GAN-based methods tend to reuse features from their training set when generating images, which, in the context of WCE image synthesis, affect the plausibility of the synthetic images (Fig. 7b) due to the limited meaningful combinations of these reused features. While GAN models have significantly contributed to the advances in medical image synthesis, especially in domains where the content is aligned within the images, such as in MRI and CT images, recent studies question their effectiveness in the context of endoscopy. This is confirmed by the results of this clinical evaluation, where VAE-based approaches outperform the GAN models in terms of realism and diversity.

It can be observed that the synthetic images produced by TIDE-II are more diverse than the real ones, and there is a narrow margin for deep generative algorithms to reach the plausibility of real images (Fig. 6). To this end, future research in deep generative methodologies for endoscopic image synthesis could focus on rethinking the role of the low-level image features, such as the color and texture, which, according to this study, seem highly correlated with the clinical relevance in endoscopic images. This investigation could improve the quality of synthetic images without compromising their clinical relevance.

In summary, this study applied the proposed CEMIS protocol for a proof-of-concept investigation assessing TIDE-II as a generative model for WCE image synthesis. The study involved 10 WCE specialists - more than in any previous study – who comprehensively evaluated endoscopy images in the context of IBD. The results indicate that TIDE-II outperforms previous GAN-based and relevant VAE-based synthesis methodologies. The findings of this study validate, from a medical perspective, the effectiveness of VAE-based methods towards generating clinically plausible endoscopy images. Considering the multilevel evaluation enabled by the proposed protocol, we believe it can serve as a reference point for future research in synthetic medical image assessment of other imaging modalities. Furthermore, the high degree of realism and diversity of the generated synthetic WCE images, renders TIDE-II a competent base model for the development of digital tools for clinicians' training, while opening perspectives for application to other medical imaging domains.

## Methods

### Clinical Protocol for Synthetic Image Assessment

This study presents a protocol that comprehensively assesses the plausibility and clinical relevance of synthetic medical images generated by AI algorithms based on clinicians' feedback. It extends previous relevant protocols[12,18] presented in the context of endoscopic image synthesis by enabling comparisons not only between synthetic and real images generated by a specific SDG method but also between different sets of images generated from different SDG methods. The proposed protocol consists of a total of five assessment procedures (*A1-A5*), organized into three parts: the first part (*A1-A3*) assesses individual images; the second part (*A4)* enables the pairwise comparison of images; the third part (*A5*) assesses the properties of groups composed of multiple images. Table 3 summarizes the tasks involved in the five assessment procedures, detailed in the following paragraphs.

### Individual Image Assessment

The first assessment procedure (*A1*) considers an equal number of real and synthetic images randomly presented one by one to clinicians without giving them any information about the images. For each image, the clinicians must complete five tasks (*A1.T1-5*). In the first task (*T1*), clinicians they have to identify if the image is real or synthetic (*T1.O1-O2*). This is followed by a second task (*T2*) that assesses the difficulty level for their choice using a five-point Likert scale (*T2.O1-O5*). T2 aims to explore the degree to which the images are correctly classified as real or synthetic and to ascertain to what extent the images synthesized by deep generative models can be considered realistic.

The third task (*T3*) aims to explain the reasons behind the clinicians' belief that the images are real or synthetic, as declared in *T1*. In the literature, the most common reasons revealing the synthetic nature of images produced by deep generative models are associated with color[13,30], blurriness[25,31], haziness and pixelation artifacts[13]. Such artifacts affect the clarity of synthetic images and consequently their overall quality. Moreover, it is usually noted that the representation of endoscopic tissues and anatomical structures in synthetic images may not be realistic enough[13,30,31], mainly due to unnatural texture or color patterns[30]. In the case of images with pathological findings, synthetic images are usually distinguished from the real ones when the appearance of lesions deviates from the expected[13,30], or when the lesions are unnaturally integrated with the background of the endoscopic images[31]. Thus, based on these considerations, suggestions *T3.O1-O5* are provided to the clinicians to explain their beliefs regarding why the images are considered unnatural. The clinicians can select one or more of the available options without any limit to the number of their choices.

**Table 3 | Overview of the proposed clinical protocol for synthetic image assessment.**

| **A1-A3 Individual Image Assessment** | **A4 Paired Image Assessment** | |
|---|---|---|
| T1: The image presented is:<br>O1: Real<br>O2: Fake | T1: Indicate which is the real image:<br>O1: Image 1<br>O2: Image 2 | |
| T2: Difficulty rate for this decision:<br>O1: Very Difficult<br>O2: Difficult<br>O3: Neutral<br>O4: Easy<br>O5: Very easy | T2: Difficulty rate for this decision:<br>O1: Very Difficult<br>O2: Difficult<br>O3: Neutral<br>O4: Easy<br>O5: Very easy | |
| T3: Reason(s) behind this decision:<br>O1: Color<br>O2: Texture<br>O3: Existence of artifacts/ luminal content<br>O4: Unrealistic appearance of anatomical structures<br>O5: Appearance of findings | T3: Reason(s) behind this decision:<br>O1: Color<br>O2: Texture<br>O3: Absence of artifacts<br>O4: Realistic anatomical structures<br>O5: Realistic appearance of findings | |
| T4: Characterize the presented image as normal or abnormal:<br>O1: Normal<br>O2: Abnormal - Erosion<br>O3: Abnormal - Erythema<br>O4: Abnormal - Ulcer<br>O5: Abnormal - Other | T4a: Characterize Image-1 as normal or abnormal:<br>O1: Normal<br>O2: Abnormal – Erosion<br>O3: Abnormal – Erythema<br>O4: Abnormal - Ulcer<br>O5: Abnormal – Other | T4b: Characterize Image-2 as normal or abnormal:<br>O1: Normal<br>O2: Abnormal – Erosion<br>O3: Abnormal - Erythema<br>O4: Abnormal - Ulcer<br>O5: Abnormal – Other |
| T5: Evaluate the quality of this image:<br>O1: Very acceptable<br>O2: Acceptable<br>O3: Moderately acceptable<br>O4: Slightly acceptable<br>O5: Not Acceptable | T5a: Evaluate the quality of Image-1:<br>O1: Very acceptable<br>O2: Acceptable<br>O3: Moderately acceptable<br>O4: Slightly acceptable<br>O5: Not acceptable | T5b: Evaluate the quality of Image-2:<br>O1: Very acceptable<br>O2: Acceptable<br>O3: Moderately acceptable<br>O4: Slightly acceptable<br>O5: Not Acceptable |
| **A5 Evaluation of Diversity and Realism** | | |
| T1: Characterize the diversity of this collection:<br>O1: Very Diverse<br>O2: Diverse<br>O3: Moderately diverse<br>O4: Slightly diverse<br>O5: Not diverse | | |
| T2: Characterize the realism of this collection:<br>O1: Very Realistic<br>O2: Realistic<br>O3: Moderately realistic<br>O4: Slightly realistic<br>O5: Not realistic | | |

The fourth task (*T4*) investigates whether the image content is normal (*T4.O1*) or abnormal (*T4.O2*), representing pathological findings. In case an abnormality is identified, the clinicians specify its type (*T4.O2.–O2.4*). The last task of this assessment procedure (*T5*) evaluates the quality of the presented images. A five-point Likert scale (*T5.O1–5*) is considered to describe the degree to which the image quality is acceptable from a clinical viewpoint.

To investigate the clinical relevance of real and synthetic images independently while establishing a baseline for the diagnostic accuracy of real images, the above assessment procedure is repeated on a random subset of images composed solely of synthetic images (assessment procedure *A2*) and on a random subset of real images (assessment procedure *A3*), without disclosing any information about the type of the images to the clinicians.

### Paired Image Assessment

The second part of the proposed protocol, implementing the fourth assessment procedure (*A4*), performs a pairwise image assessment. Pairs of real and synthetic images are randomly presented to the clinicians for a side-by-side comparison. This strategy is designed to mitigate any potential expectation bias in the previous image assessment procedures (*A1-A3*). Expectation bias can be attributed to preconceived notions regarding the quality of real and synthetic images. Thus, a direct comparison between them is conducted, with the evaluation emphasizing details in features and structures illustrated, rendering the synthetic and real images distinguishable.

Image pairs are presented to the clinicians for examination. For each image pair, the clinicians have to complete five tasks (*A4.T1-T5*). The first task (*A4.T1*) assesses which image is the real one. The second task (*A4.T2*) aims to assess the confidence of that choice. A five-point Likert scale is employed for this purpose. The third task (*A4.T3*) aims to justify the selection of the clinicians in task *A4.T2*, based on the image features used in the third task of the previous assessment procedures (*A1-A3.T3*). Also, similar to the previous procedures, clinicians provide their opinions regarding the presence of any pathological findings (*A4.T4*) and assess the quality of the images presented (*A4.T5*).

### Evaluation of Diversity and Realism

This is the last part of the proposed protocol, implementing the fifth assessment procedure (*A5*). It aims to assess the diversity of the synthetic images and the degree to which they look real, by providing them group-wise to the clinicians. To this end, randomly sampled groups of images are generated using various generative models, The image groups are provided to the experts to assess their diversity (*A5.T1-T4a*) and realism (*A5.T1-T4b*). Likert scales of five-points included in Table 3 are used to capture the clinicians' feedback. In both cases, the medical experts select one of the five options provided.

The following paragraphs provide a brief but comprehensive background on state-of-the-art generative models, used in the comparative evaluations performed by this study, as well as the architectural details of the proposed TIDE-II model.

### Generative Adversarial Networks - GANs

A GAN model[4] consists of two neural networks a Generative network which is called Generator and a Discriminative network which is called Discriminator. The Generator learns to synthesize realistic data samples drawn from a real data distribution, while the Discriminator tries to distinguish between real and synthetic samples. The training procedure of a GAN model can be represented via a min-max game where the training parameters of the two networks are optimized concurrently through an adversarial process. During this procedure the Discriminator network is trained to maximize the probability of correctly identifying real and fake samples, while the Generator network is trained to minimize the probability with which the Discriminator correctly identifies generated samples as fake. The optimization process of a GAN model is formulated as a min-max game, with objective function *V(D, G)*, which is described by the following equation:

$$\min_G \max_D V(D,G) = \mathbb{E}_{x \sim p(x)_{data}}[\log D(x)] + \mathbb{E}_{z \sim p(z)}[\log(1 - D(G(z)))] \quad (1)$$

where, $\mathbb{E}$ represents the expected value, x is a real data sample derived from a distribution of real data $p(x)_{data}$ , $z$ is a noise vector sampled from a prior Gaussian or uniform distribution $p(z)$. $G(z)$ denotes the synthetic sample produced by Generator network, $D(x)$ denotes the probability that sample x derives from the real data distribution and $D(G(z))$ denotes the probability that the synthetic data sample belongs to the distribution of real data.

In fact, most existing methods for endoscopic image generation focus on GANs. In our previous research work[26], the first study in the context of endoscopic image synthesis, we presented a non-stationary Texture Synthesis GAN (TS-GAN) model for generating images with and without inflammation conditions. TS-GAN model is trained to synthesize endoscopic images based on randomly selected texture patches extracted from real images. Following research works employed various GAN frameworks to generate endoscopic images. Among them CycleGAN[32] and variants of StyleGAN models [19,33] are considered the most prevalent methods in the literature. Although, these architectures have been proposed for synthesizing non-medical images they consist common approaches for endoscopic image synthesis[11,14]. CycleGAN model has been proposed in the context of unpaired image-to-image translation between two domains i.e. a source domain (usually normal endoscopic images) and a target domain (usually abnormal endoscopic images). CycleGAN consists of two GAN models, each corresponding to one domain trained simultaneously through a single loss function that ensures that mappings learned between two domains are consistent. StyleGAN model modifies the traditional GAN architecture by incorporating a style-based method. It employs an FCN network trained simultaneously with the GAN model to learn style mappings corresponding to the latent representations of real data. These mappings capture various visual features such as textures, colors, and patterns associated to different concepts in real images. The learned style vectors guide the image synthesis process of the generator network through a normalization technique, called Adaptive Instance Normalization (AdaIN). By adjusting the style parameters associated with different layers of the FCN during inference procedure, a conditional manipulation of the attributes underlying the real data distribution is achieved, over the generated images. Therefore, considering the broad applicability of the GAN models for endoscopic image synthesis, this study includes comparative results using CycleGAN, StyleGANv2, and TS-GAN models.

### Variational Autoencoders - VAEs

A VAE[5] extends the conventional autoencoder (AE) model by introducing probabilistic modeling. A typical AE consists of two parts: the encoder network and the decoder network. The Encoder network maps an input volume to a compressed representation, known as latent space representation, while the decoder network reconstructs the initial data volume based on the representation of the latent space, by minimizing a loss function quantifying the image reconstruction error. However, in AEs input volumes are encoded into specific, fixed representations in the latent space, which subsequently are used to reconstruct the corresponding output volumes. These deterministic encodings estimated in AEs affect the regularity and interpretability of the latent representations, since fixed mappings often result in less smooth latent spaces, which adversely impacts the representation of complex data distributions. In a VAE model the latent space is regarded as a probability distribution and thus the typical AE is extended twofold: by employing variational inference to approximate the posterior distribution of the latent space; and by regularizing the approximate posterior distribution measuring its Kullback–Leibler (KL) divergence from a prior distribution. Assuming that the true posterior distribution is intractable, an efficient approximation of the actual posterior distribution can be performed by maximizing a variational lower bound according to the following equation:

$$\mathscr{L}(\boldsymbol{\vartheta}, \boldsymbol{\varphi}; \boldsymbol{x}_i) = -KL\,(q\,(\boldsymbol{z}|\boldsymbol{x}_i; \boldsymbol{\varphi}) \,\|\, p\,(\boldsymbol{z}; \boldsymbol{\vartheta})) + \mathbb{E}_{q\,(z\,|x_i;\,\varphi)} \log p(\boldsymbol{x}_i|\boldsymbol{z}\,;\boldsymbol{\vartheta}) \quad (2)$$

where, $x_i$ denotes the training samples, $z$ denotes the latent variables, $q(\boldsymbol{z}|\boldsymbol{x};\boldsymbol{\varphi})$ is the approximated posterior distribution estimated by the encoder network, $p(\boldsymbol{z};\boldsymbol{\vartheta})$ is a prior distribution, $\mathbb{E}q(\boldsymbol{z}\,|\boldsymbol{x}_i;\,\boldsymbol{\varphi})$ denotes the expected value with respect to the approximate posterior distribution $q(\boldsymbol{z}|\boldsymbol{x}i;\,\boldsymbol{\varphi})$, log $p(\boldsymbol{x}i|\boldsymbol{z}\,;\boldsymbol{\vartheta})$ is the log-likelihood of the observed data $\boldsymbol{x}_i$ given the latent variable $\boldsymbol{z}$, and $\boldsymbol{\varphi}$, $\boldsymbol{\vartheta}$ represent the training parameters of the encoder and decoder networks respectively. Conventionally, the

prior distribution $p(\boldsymbol{z};\boldsymbol{\vartheta})$ is formulated as a multivariate Gaussian distribution $\mathcal{N}(\boldsymbol{z};0,\mathbf{I})$. Letting the true intractable posterior distribution $p(\boldsymbol{z}|\boldsymbol{x}i\ ;\boldsymbol{\vartheta})$ be an approximation of the Gaussian with an approximately diagonal covariance estimated according to Eq. (3):

$$\log q\ (\boldsymbol{z}\ |\boldsymbol{x}_i;\ \boldsymbol{\varphi}) = \log \mathcal{N}\ (\mathbf{z};\ \boldsymbol{\mu}_i\ ,\ \boldsymbol{\sigma}_i^2\ \mathbf{I}) \tag{3}$$

where $\boldsymbol{\mu}_i$ and $\boldsymbol{\sigma}^2_i$ represent the distribution parameters estimated by encoder network

Eq. (1) is formulated as follows:

$$\mathcal{L}(\vartheta, \varphi;\ \mathbf{x}_i) \simeq \frac{1}{2}\sum_{j=1}^{J}\left(1 + \log \boldsymbol{\sigma}_{ij}^2 - \boldsymbol{\sigma}_{ij}^2 - \boldsymbol{\mu}_{ij}^2\right) + \frac{1}{L}\sum_{l=1}^{L} \log \mathrm{p}(\mathbf{xi}|\mathbf{z}_l\ ;\vartheta) \tag{4}$$

where, $z_l = \boldsymbol{\mu}_i + \boldsymbol{\sigma}_i \odot \epsilon_l$ and $\epsilon_l \sim \mathcal{N}\ (0, \mathbf{I})$, J refers to the dimensionality of the underlying manifold, L refers to the sample size of the Monte Carlo method sampling from the approximate posterior distribution of the encoder, and symbol $\odot$ represents the Hadamard product operation.

Recent studies [13,25] in WCE image synthesis have presented VAE-based models in the context of generating high-quality images containing inflammation conditions. Our previous study, presented a Convolutional VAE model, namely EndoVAE[25], that incorporates residual connections towards improving the variety of the synthetic images over the GAN-based approaches. Subsequently, our following study expanded this architecture with a multiscale feature extraction scheme aiming to preserve more details in synthesis results. This later model proposed, namely TIDE[13] , was designed to capture feature-rich representations of the endoscopic tissue structures, that contain information at different scales enabling the generation of images with improved variety and finer details.

### TIDE-II: Revisiting Model Architecture for Improved Image Synthesis

This study uses the proposed protocol to clinically evaluate the image synthesis performance of a novel VAE model, named TIDE-II. Unlike the current relevant models, TIDE-II generates high quality, high resolution WCE images, which are suitable for clinical use, *e.g.*, clinicians' training in WCE reviewing. This is achieved by enhancing our state-of-the-art TIDE model by incorporating components of the ConvNeXt[34] CNN architecture.

The architecture of TIDE-II can be broken down into two parts: the encoder and the decoder. The encoder aims to encode an input image by extracting features using consecutive down-sampling layers, leading to a latent representation of the input volume, while the decoder part uses consecutive up-sampling layers to synthesize high resolution images. The encoder part of TIDE-II model is a modified version of ConvNeXt-tiny[34] model, which incorporates several modern techniques inspired by Visual Transformer (ViT)[27], enhancing the feature extraction capabilities of conventional CNN architectures. More specifically, the encoder consists of four down-sampling operations each followed by a varying number of ConvNeXt blocks. Down-sampling is performed using a convolution layer followed by output normalization. The ConvNeXt blocks are composed by a depth-wise separable convolution[35] followed by output normalization and point-wise convolution which linearly combines the extracted features across the depth axis. Opposed to conventional Rectified Linear Unit (ReLU)[36] activations, ConvNeXt block uses Gaussian Error Linear Units (GELU)[37] which provide smoother activations benefiting the gradient flow while training. Towards that, residual connections[38] are also used, mitigating the vanishing gradient problem. While output normalization is an effective tool in model training regularization, findings from our previous work[13] show that it can lead to unstable training in VAE architectures resulting to exploding or vanishing KL divergence. As a result, TIDE-II architecture removes the output normalization operations, which contributes to a more stable training of the model. The modified version of ConvNeXt block is illustrated in Supplementary Fig. 9.

After the encoder, the extracted features are flattened to enter a shallow fully connected network, composed by one fully connected layer followed by two parallel fully connected layers, which are used to estimate the parameters $\mu$, $\sigma$ of the latent space distribution respectively. Random sampling from a standard normal distribution $\mathcal{N}$ (0, $\mathbf{I}$) is performed in the latent space. Based on the distribution parameters estimated by the network, a latent volume is sampled using the reparameterization trick according to Eq. (4). This sampled latent volume is then forwarded to the decoder part of the architecture.

The decoder part of the architecture receives a vector from the latent space, which using a fully connected layer gets reconstructed into the feature space of the encoder. Similarly to the encoder, the decoder uses a series of modified ConvNeXt blocks and up-sampling operations using transposed convolution layers to reconstruct the initial input volume. In addition to the encoder, the decoder part of the architecture makes use of a Multi-Scale feature extraction Block (MSB)[13] on the last transposed convolution operation. The MSB module extracts features from multiple scales aiming to capture different levels of detail from the input feature volume, enhancing the overall quality of the reconstruction process and introducing a higher degree of diversity of the results. The architecture of the MSB is illustrated in Supplementary Fig. 10. The last layer of the decoder is a convolutional layer which generates the output of the model. The architecture of the entire TIDE-II model is provided in Supplementary Fig. 11.

## Author contributions

P.G., D.E.D, and D.K.I. contributed to the design of this study, the development of the proposed protocol and the generative model, P.G. improved the original model, performed the experiments, statistically analyzed the results, and wrote the initial draft with the assistance of D.E.D and D.K.I.. A.K. contributed to medical aspects of the design of the study. A.K. and the rest of the authors participated as experts, contributing to the evaluation of the results. All authors contributed to the proofreading and correction of the final manuscript and agreed to its final version and its submission for publication.

## Competing interests

A.K. reports that he is a co-director and shareholder of AJM Medicaps, and shareholder of iCERV Ltd; and has received consultancy fees (Jinshan Ltd. and DiagMed Healthcare Ltd.), travel support (Jinshan DiagMed Healthcare Ltd., Aquilant), research support (grant) from ESGE/Given Imaging Ltd. and (material) IntroMedic/SynMed, lecture honoraria (Jinshan, Medtronic). He participated in advisory board meetings for Tillots, ANKON, Dr Falk Pharma UK. The rest of the authors declare no competing interests.

## Additional information

**Supplementary information.** The online version contains supplementary material available at *'url to be included after acceptance'*.

## Data Availability

All datasets used in this study are publicly available. KID is available at: https://mdss.uth.gr/datasets/endoscopy/ kid/. Kvasir is available at: https://datasets.simula.no/ kvasir/.

**Correspondence** and requests for materials should be addressed to Dimitris K. Iakovidis

## Supplementary Information

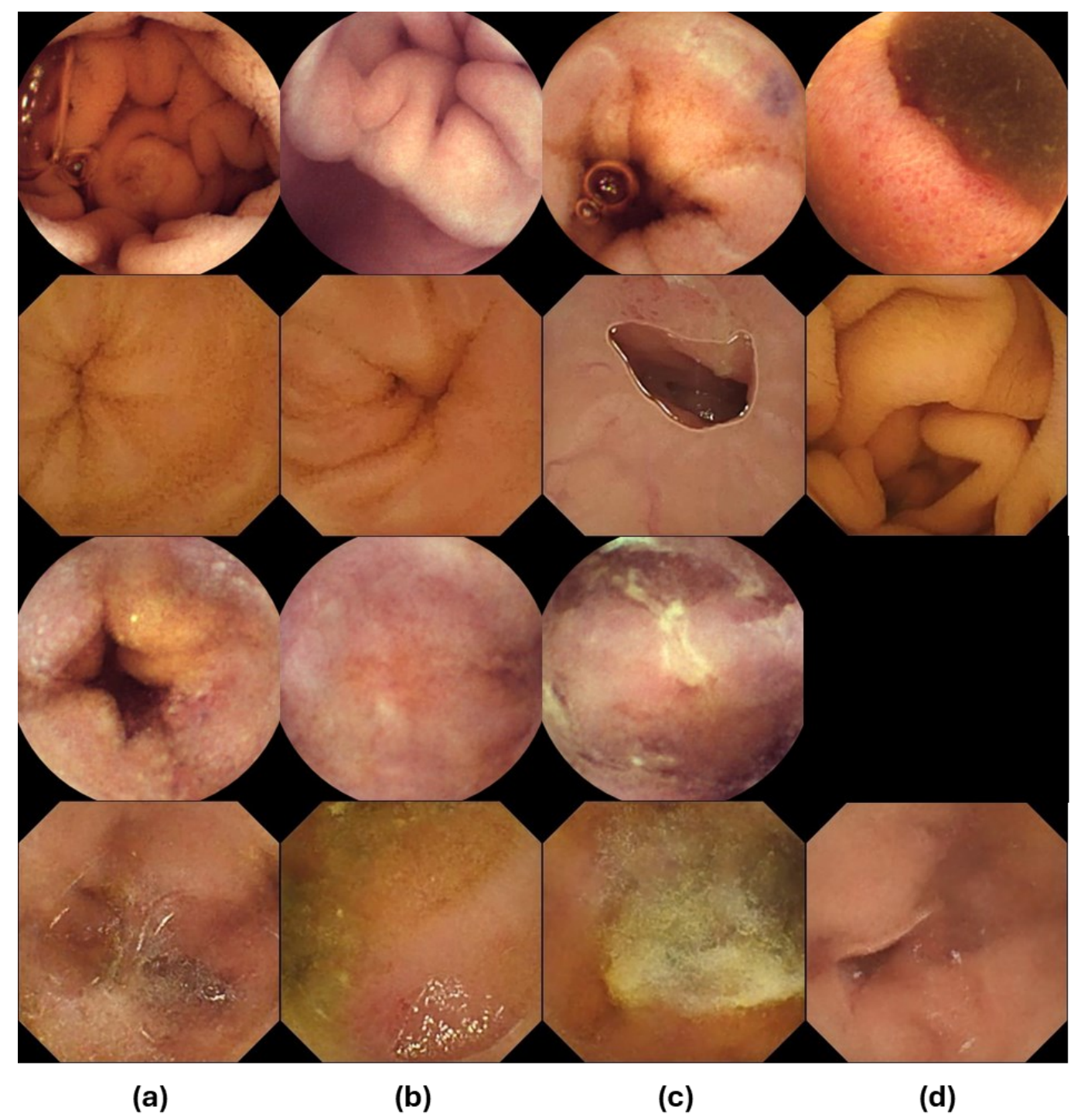

**Supplementary Figure 1 | Correctly classified images by all medical experts in assessment procedure A1, per experience group.** The medical experts were requested to classify the images as real or synthetic. The first and second rows illustrate correctly classified real images from the KID and Kvasir datasets, respectively. The third and fourth rows illustrate correctly classified synthetic images generated by TIDE-II, from KID and Kvasir datasets, respectively. The images are ordered in columns according to the experts' years of experience in WCE reviewing. Images correctly classified by all experts with **(a)** 5-10 years, **(b)** 10-20 years, **(c)** over 20 years of experience, and **(d)** all experts; the gap indicates that none of the images generated by TIDE-II based on the KID dataset was correctly classified as synthetic by all experts.

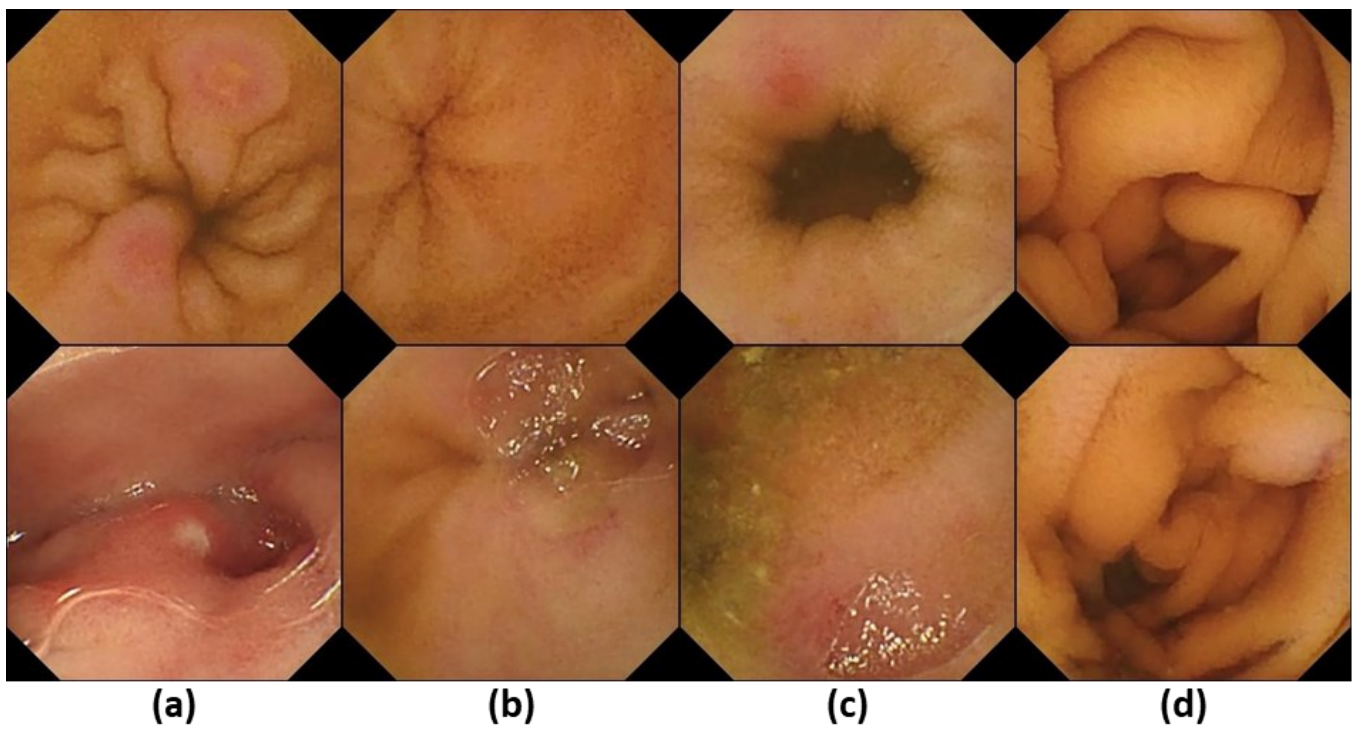

**Supplementary Figure 2 | Representative images rated with acceptable image quality by the medical experts in assessment procedure A1.** The first row illustrates real images, and the second illustrates synthetic images generated by TIDE-II. Experts with **(a)** 5-10, (b) 10-20 years, and **(c)** over 20 years of experience in capsule endoscopy. **(d)** All the groups of medical experts.

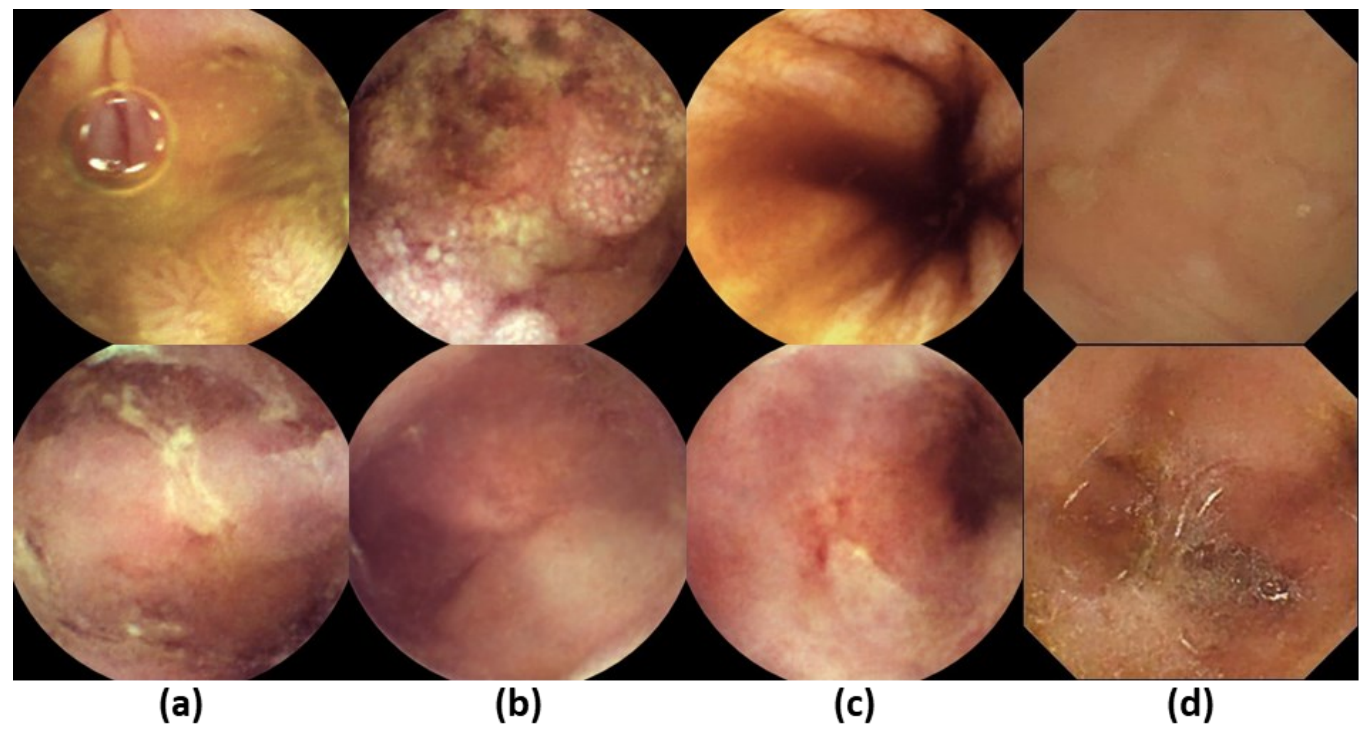

**Supplementary Figure 3 | Representative images rated with slightly acceptable image quality by the medical experts in assessment procedure A1.** The first row illustrates real images, and the second illustrates synthetic images generated by TIDE-II. Experts with **(a)** 5-10, (b) 10-20 years, and **(c)** over 20 years of experience in capsule endoscopy. **(d)** All the groups of medical experts.

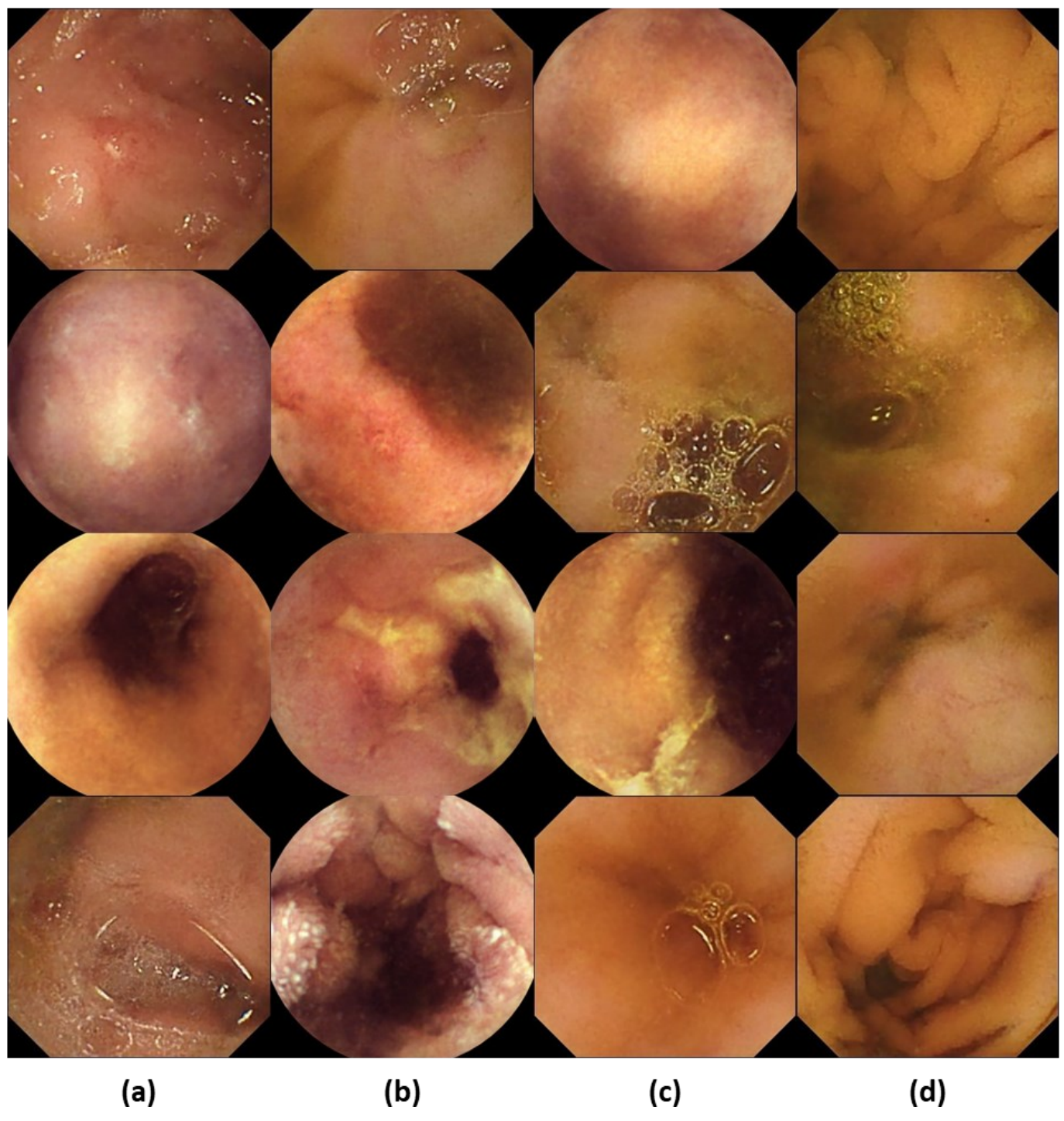

**Supplementary Figure 4 | Synthetic images generated by TIDE-II, misperceived as real by all the medical experts in assessment procedure A2, per experience group.** (a) Synthetic images misperceived as real by the medical experts participated in the assessment. Experts with (a) 5-10 years (b) 10-20 (c) over 20 years of experience in capsule endoscopy and (d) all the experts.

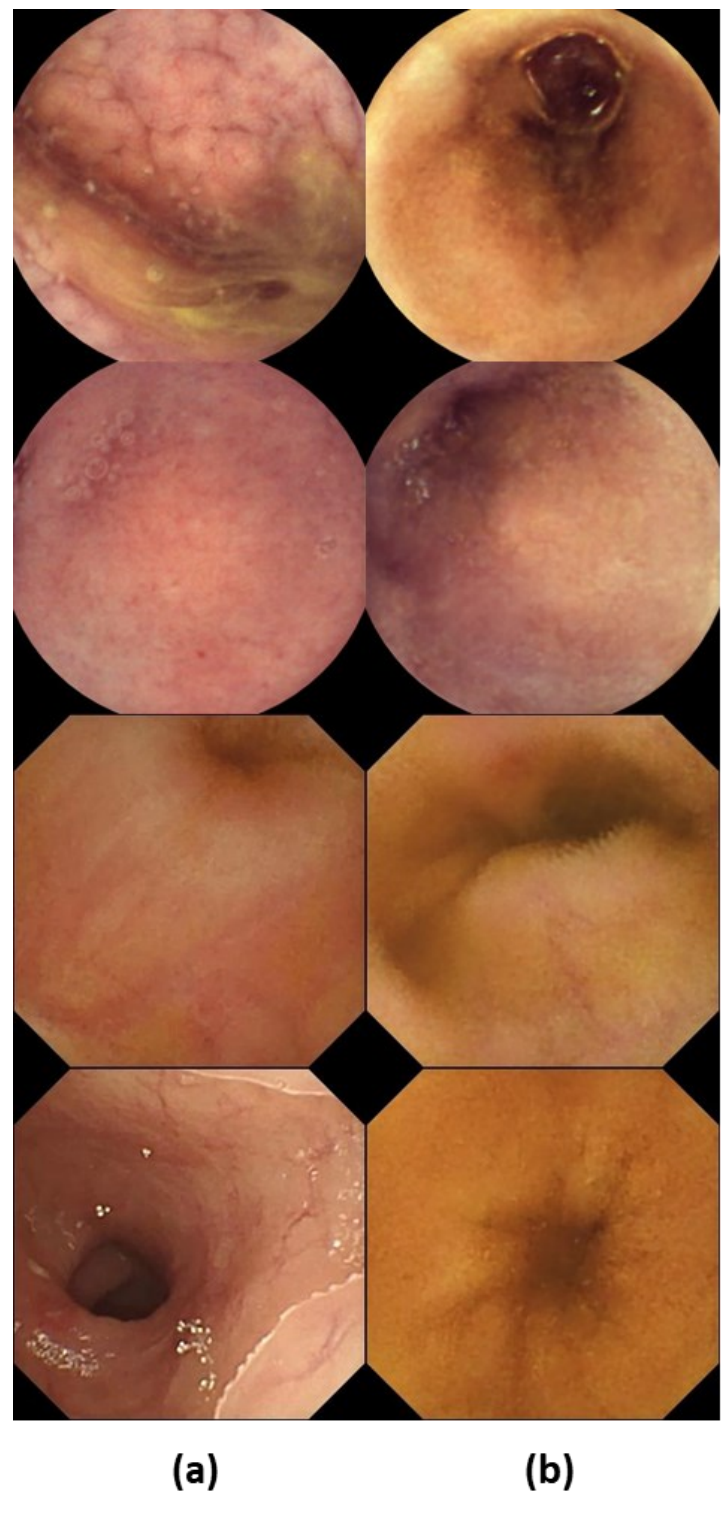

**Supplementary Figure 5 | Most mispredicted pairs of images in assessment procedure A4.** Image pairs in which the experts most frequently mispredicted the real image. These are also the pairs that the experts rated more frequently as difficult and very difficult. **(a)** Real images. **(b)** Synthetic images.

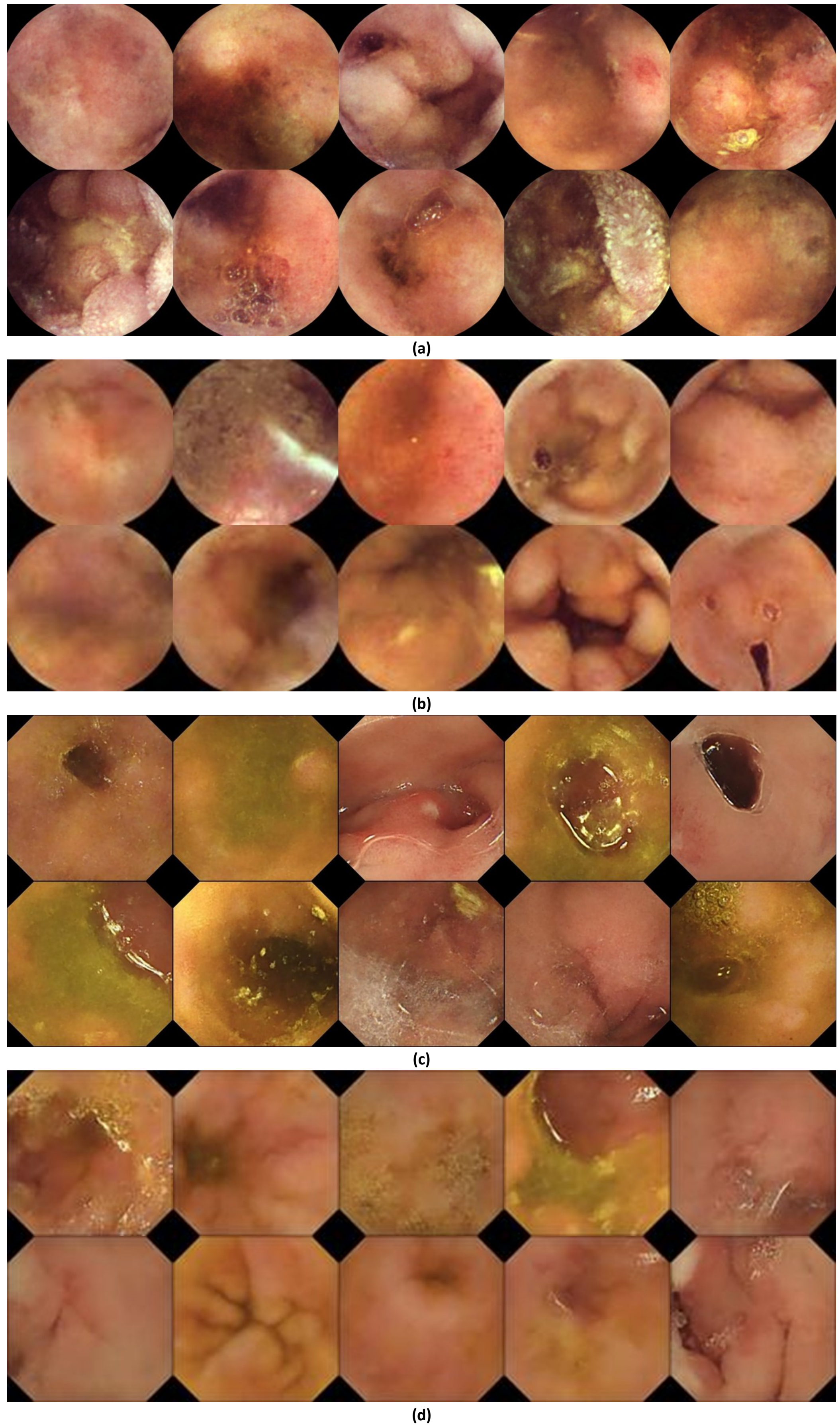

**Figure 6 | Most realistic image collections, as rated by the experts in the assessment procedure A5**. **(a)** Images generated by TIDE-II based on KID. **(b)** Images generated by TIDE based on KID. **(c)** Images generated by TIDE-II based on Kvasir-Capsule dataset. **(d)** Images generated by TIDE based on Kvasir-Capsule dataset.

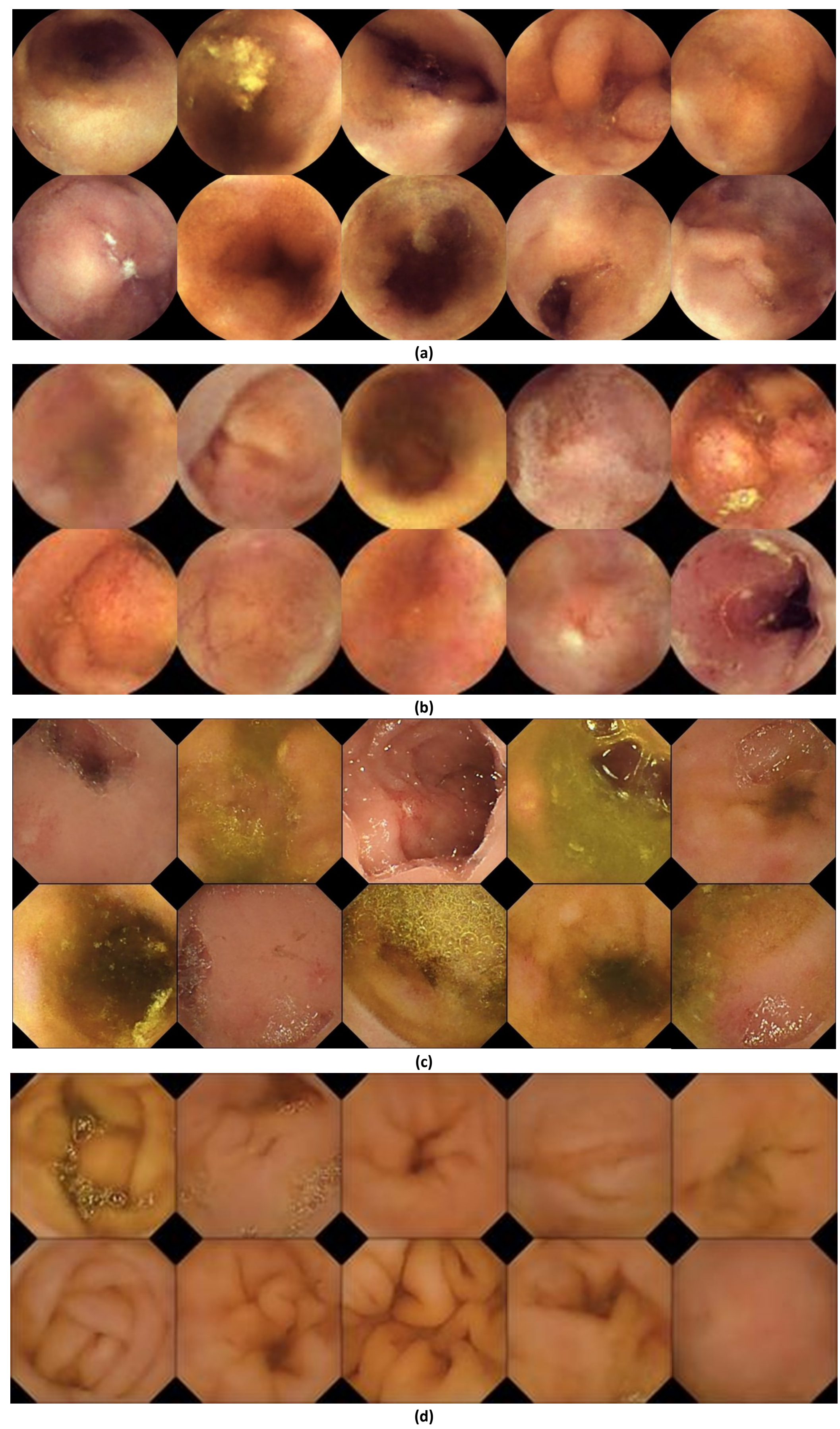

**Supplementary Figure 7 | Most diverse image collections, as rated by the experts in the assessment procedure A5**. **(a)** Images generated by TIDE-II based on KID, **(b)** Images generated by TIDE based on KID. **(c)** Images generated by TIDE-II based on Kvasir-Capsule dataset. **(d)** Images generated by TIDE based on Kvasir-Capsule dataset.

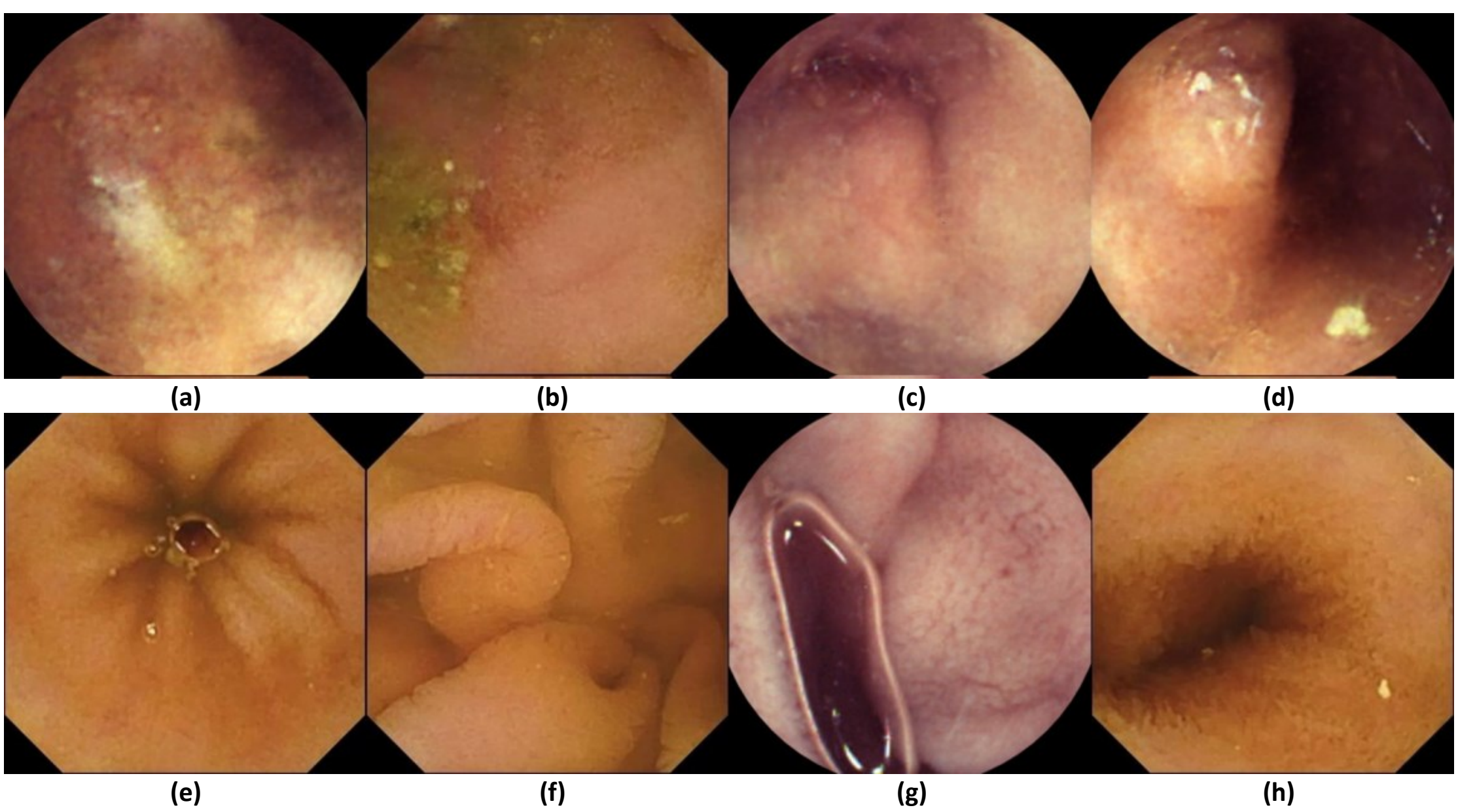

**Supplementary Figure 8 | Images associated with the existence of artifacts in individual image assessments A2 and A3**. In individual image assessments A2 and A3 medical experts evaluated solely synthetic and solely real images, respectively. The first row illustrates synthetic images containing artifacts from assessment A2 and the second row illustrates real images containing artifacts from assessment A3 according to the experts' opinions.

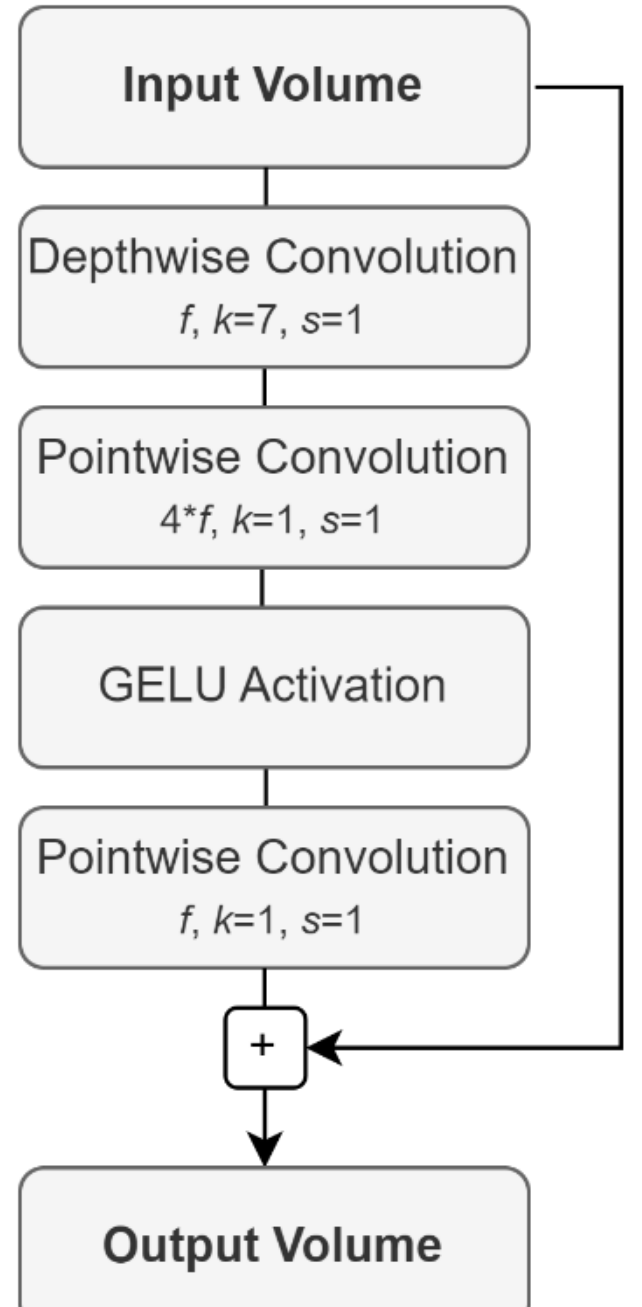

**Supplementary Figure 9 | Modified ConvNeXt Block used in TIDE-II architecture**. Symbol *f* denotes the number of filters used in each convolution layer. Symbol *k* denotes the kernel size and symbol *s* denotes the stride of convolution operation. The number of filters *f* varies for the ConvNeXt blocks interposed between the downsampling operations in the Encoder part and the upsampling operations in the Decoder part of TIDE-II architecture. In the Encoder network, it is equal to 96 for the first three ConvNeXt blocks, 192 for the next three ConvNeXt blocks, 384 for the next nine ConvNext blocks and 784 for the last three ConvNeXt blocks. In the Decoder Network the number of filters *f* is symmetric to the Encoder, yet with a reverse order.

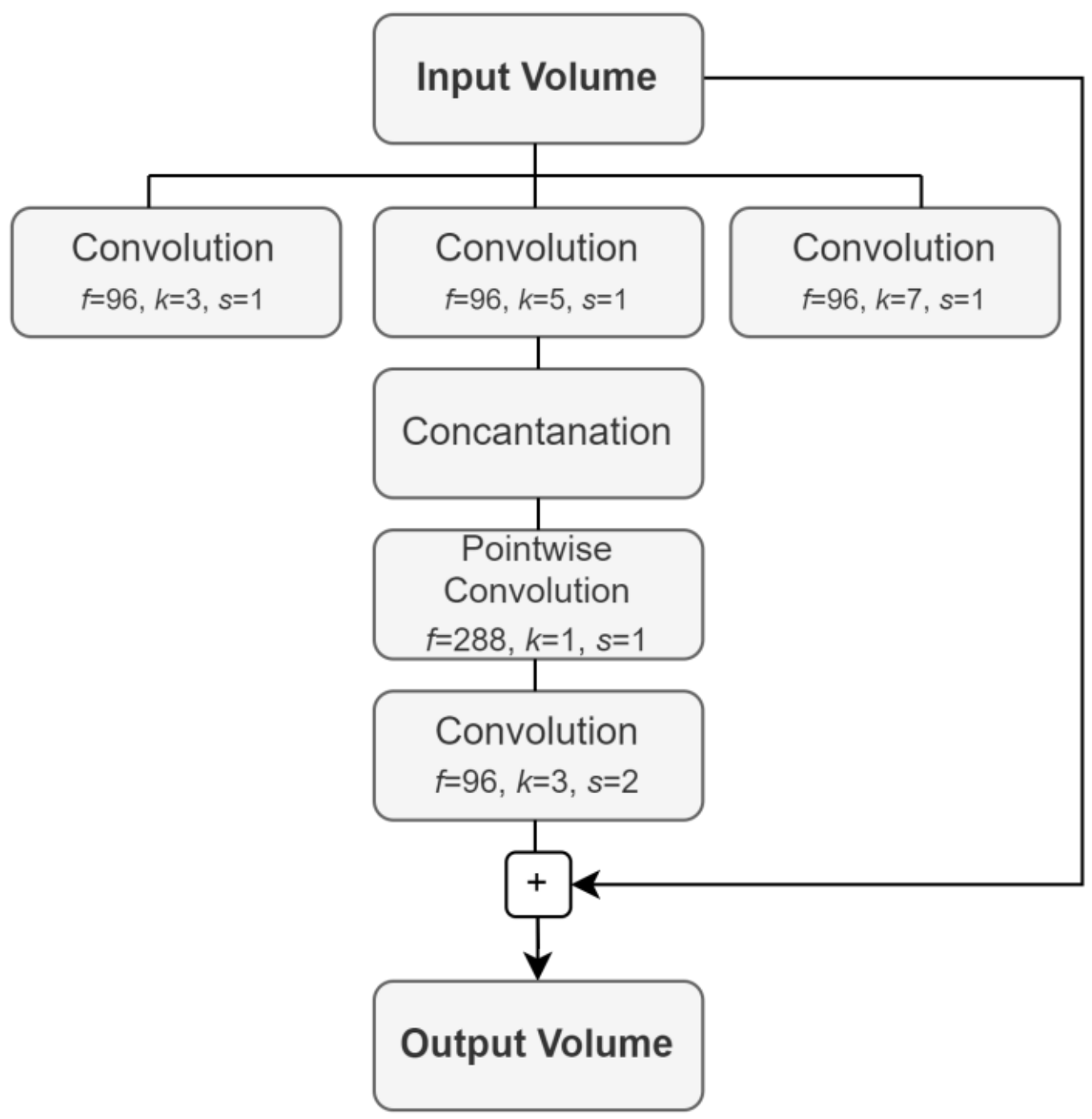

**Supplementary Figure 10 | Architecture of Multi Scale feature Extraction Block (MSB) used in TIDE-II architecture**. Symbol *f* denotes the number of filters, symbol *k* denotes the kernel size and symbol *s* denotes the stride of convolution operations.

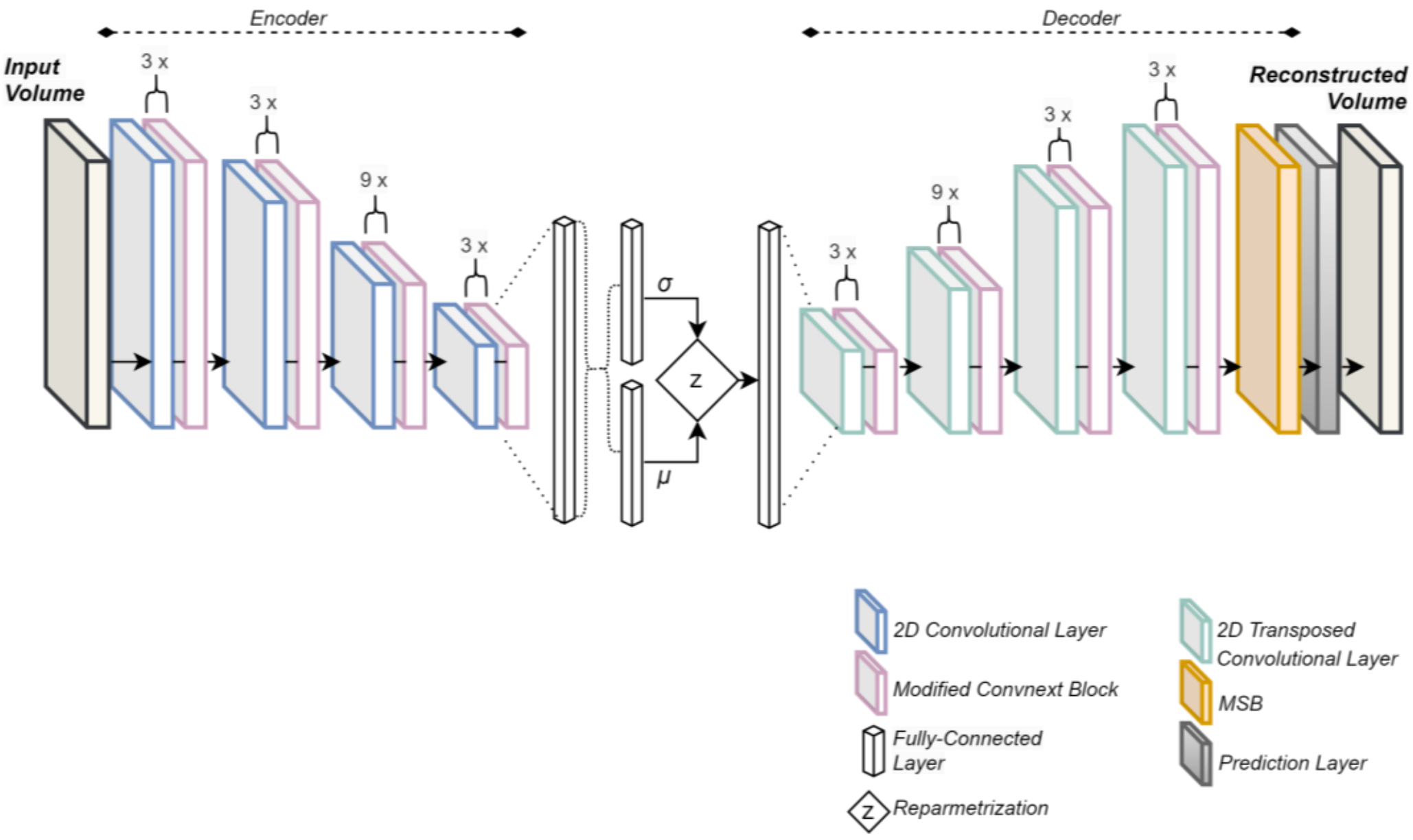

**Supplementary Figure 11 | Architecture of the proposed TIDE-II model.**